\pdfoutput=1

\documentclass[11pt]{article}

\usepackage{acl}

\usepackage{times}
\usepackage{latexsym}

\usepackage[T1]{fontenc}

\usepackage[utf8]{inputenc}

\usepackage{microtype}

\usepackage{inconsolata}

\usepackage{amsmath}
\usepackage{fontawesome}
\usepackage{amssymb}
\usepackage{mathtools}
\usepackage{amsthm}
\usepackage{hyperref}
\usepackage{url}
\usepackage{caption}
\usepackage{wrapfig}
\usepackage{booktabs}
\usepackage{multirow}
\usepackage{diagbox}
\usepackage{graphicx}
\usepackage{natbib}
\newcommand{\STAB}[1]{\begin{tabular}{@{}c@{}}#1\end{tabular}}

\title{Low-Rank Interconnected Adaptation across Layers}

\author{
 \textbf{Yibo Zhong\textsuperscript{1}},
 \textbf{Jinman Zhao\textsuperscript{2}},
 \textbf{Yao Zhou\textsuperscript{1}}\thanks{Corresponding author}\thanks{This work was supported by National Natural Science Foundation of China (Grant 62376172)},
\\
\\
 \textsuperscript{1}Sichuan University,
 \textsuperscript{2}University of Toronto,
\\
   \href{}{\texttt{zhongyibo@stu.scu.edu.cn}} \quad \href{mailto:yaozhou@scu.edu.cn}{\texttt{yaozhou@scu.edu.cn}}
}

\begin{document}
\maketitle
\begin{abstract}
Low-rank adaptation (LoRA) is a widely used parameter-efficient fine-tuning (PEFT) method that learns weight updates $\Delta W = AB$ for pretrained weights $W$ through low-rank adapters $A$ and $B$. While LoRA ensures hardware efficiency, its low-rank weight updates limit adaptation performance. In this paper, we propose \underline{l}ow-rank \underline{i}nterconnected adaptation across \underline{l}a\underline{y}ers (Lily), a novel PEFT method that introduces an interconnected framework with locally shared $A$ and globally shared $B$ experts. This structure eliminates redundant per-layer $AB$ pairs, enabling higher-rank $\Delta W$ with equal or fewer parameters. To enhance expressiveness, we use data-dependent routers to determine $A$-$B$ interconnections, preventing $B$ experts from converging to the same behavior and improving representational power across domains. Experiments across modalities, architectures, and model sizes demonstrate Lily's superior performance and efficiency. \href{https://github.com/yibozhong/lily}{\faGithub~Github}
\end{abstract}

\section{Introduction}

\begin{figure}[h]
    \centering
    \includegraphics[width=0.99\linewidth]{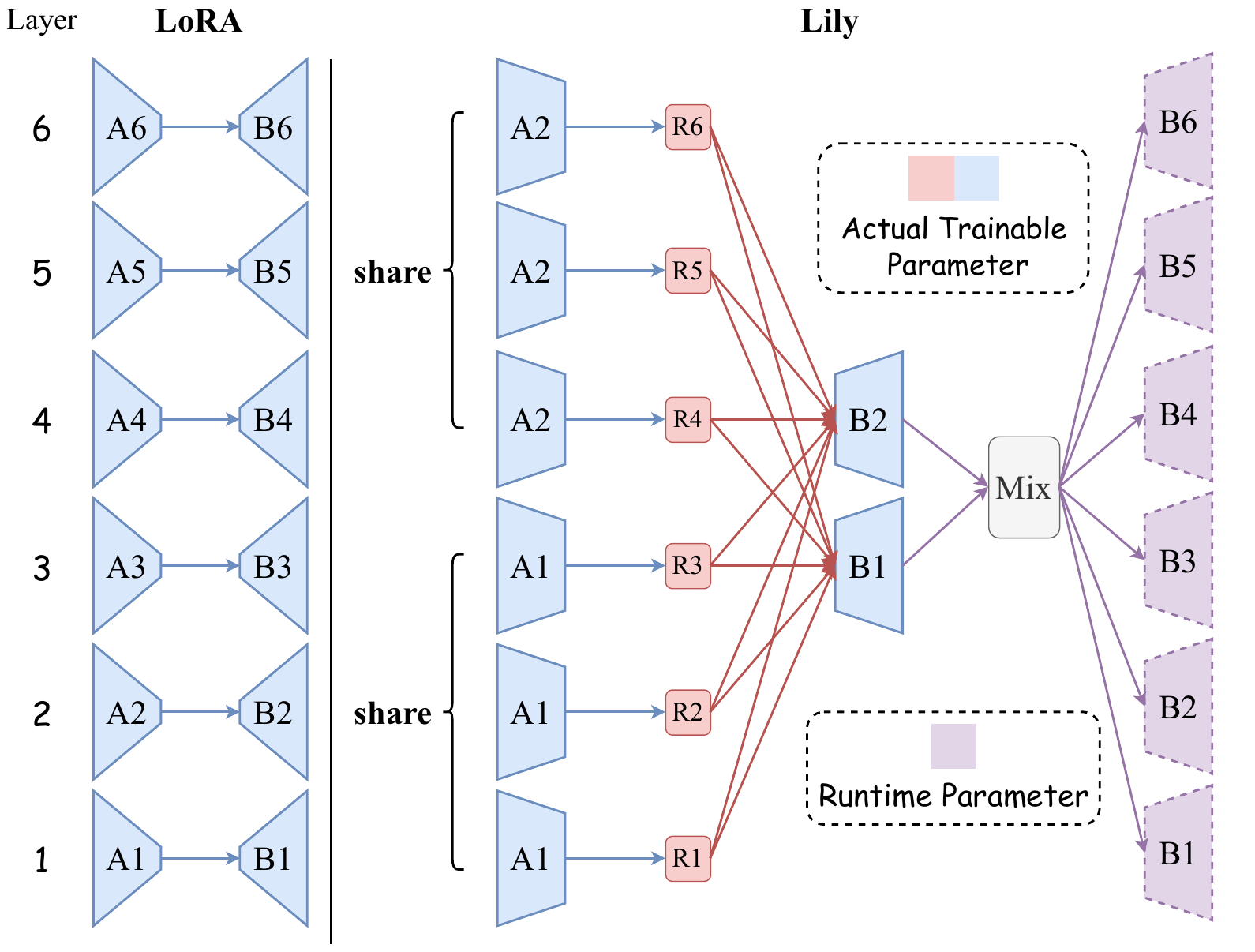}
    \vspace{-20pt}
    \caption{Dynamics of LoRA and Lily. In this 6-layer example with a fixed overall parameter budget, LoRA allocates the same parameter budget to each layer, resulting in small rank updates for the weights. Lily overcomes this by employing a small number of shared adapters with a much larger rank, achieving higher-rank updates while using the same or even a smaller parameter budget. Considering the different characteristics, and to make the adaptation more dynamic, the adapters are mixed according to a data-dependent router, represented by $R$.}
    \label{fig:cmp-of-lora}
    \vspace{-15pt}
\end{figure}

Fine-tuning foundation models like Transformers \citep{vaswani2017transformer} on downstream tasks is common but costly, especially for large models like LLMs, which incur high computational and storage demands and risk catastrophic forgetting \citep{biderman2024lora_forget}. Linear probing alleviates these issues by fine-tuning only the final modules, but suffers from performance loss due to frozen backbone weights. To address this, parameter-efficient fine-tuning (PEFT) freezes the backbone and introduces lightweight modules for task-specific learning. Among PEFT methods, Low-rank Adaptation (LoRA \citep{hu2021lora}) is widely used, particularly for LLMs. LoRA introduces low-rank projection matrices, $A$ and $B$, to approximate weight updates $\Delta W$, achieving significant savings in computation and storage while outperforming linear probing by updating the backbone weights.

However, LoRA and its subsequent improvements \citep{miles2024velora, zhang2023adalora, zhong2024convlora} face a limitation: the learned weight updates $\Delta W$ are constrained to be low-rank, limiting model performance. A key issue is that LoRA allocates the same parameter budget to each layer, regardless of their importance (Fig. \ref{fig:cmp-of-lora}). As a result, the rank of each adapter is constrained by the fixed budget, raising the critical question: \textit{Can we enable more dynamic, expressive adaptation with high-rank weight updates under the same parameter budget?}

In this paper, we propose \underline{\textbf{L}}ow-rank \underline{\textbf{i}}nterconnected adaptation across \underline{\textbf{l}}a\underline{\textbf{y}}ers (Lily), a novel framework for more expressive and efficient PEFT. Specifically, we decouple $A$ and its corresponding upward $B$, eliminating their tight coupling. Each $A$ is connected to all $B$s, and vice versa, as illustrated in Fig. \ref{fig:cmp-of-lora}. This creates a hierarchical structure where locally-shared $A$s perform downward projections at specific layers, while globally-shared $B$s perform upward projections across all layers. To enhance dynamism, we selectively connect each $A$ with $B$s based on layer features. $A$ extracts features from the current layer, and a selective mixture of $B$s is performed based on these features, enabled by routers \citep{shazeer2017MoE} that generate data-dependent weight distributions for $B$ experts.

The interconnected structure makes the adaptation process more dynamic and flexible, with rich interactions between adapters. By reducing the number of adapters and increasing their rank, \textbf{Lily achieves higher-rank weight updates than LoRA while using the same or fewer parameters.} Additionally, Lily enables comprehensive information access and learning by allowing adapters at each layer to collaborate, share knowledge, and model dependencies across layers. Our key contributions include:

\begin{itemize}
    \item We propose Lily, a novel PEFT framework that introduces interconnected adapters, effectively overcoming the limitations of low-rank weight updates in LoRA under the same parameter constraints.
    \item Lily utilizes routers to dynamically select and connect an adapter $A$ with multiple adapter $B$ experts, enabling richer information flow and more expressive adaptation dynamics.
    \item Extensive experiments are conducted across diverse modalities, architectures, and model scales, demonstrating Lily's superior performance and efficiency in a wide range of scenarios.
\end{itemize}

\section{Related Work}

\textbf{Parameter Efficient Fine-Tuning} Foundation models are typically pre-trained on large datasets and fine-tuned on downstream tasks. Parameter-efficient fine-tuning (PEFT) seeks to fine-tune models efficiently with minimal parameters while maintaining performance and preserving learned knowledge. It effectively addresses limitations of conventional fine-tuning techniques, like full fine-tuning or linear probing. Current PEFT approaches can be divided into two categories: 1) adapter-based methods \citep{hu2021lora,chen2022adaptformer,pfeiffer2020adapter-P,jie2023fact,houlsby2019parameterEfficientTLforNLP, zhong2025heartlora}
 and 2) prompt-based methods \citep{tu2023VQT, tu2023VPT}. Adapter-based methods insert lightweight adapters into the Multi-Head Self-Attention (MHSA) or Feed-Forward Network (FFN) blocks of the Transformer architecture, while prompt-based methods add trainable tokens to the input sequence.

Among these, low-rank adaptation (LoRA \citep{hu2021lora}) is a well-known technique. It introduces projection matrices $A$ and $B$ for each adaptation target $W$, where $A$ projects input $x$ to a low-dimensional space and $B$ restores it to the original dimension. The product of these matrices approximates the weight update $\Delta W$ in full fine-tuning (FFT). However, this limits the update to a low-rank subspace, which may affect performance. Additionally, $A$ and $B$ are tightly coupled, restricting the adaptation process to information from the current layer, which may hinder the modeling of dependencies across layers.

\textbf{Mixture of Experts} Mixture of Experts (MoE) is an active research area that has received significant attention, especially in the field of large language models (LLMs). Conditional computation, which activates different parts of the network on a per-example basis, has been proposed to enhance model capability without increasing computation \citep{davis2013condition1,bengio2013condition2,eigen2013condition3,almahairicondition4}
. The sparsely-gated MoE layer is introduced to implement this idea, consisting of numerous sub-networks \citep{shazeer2017MoE}. A trainable gating network (router) determines the combination of experts for each example. There are already PEFT methods like MoLORA \citep{zadouri2023MoLORA} and MOLA \citep{gao2024MOLA} that apply the MoE design concept to PEFT. However, these methods simply treat the adapters $A$ and $B$ combined in LoRA as a single expert. Concurrent research \citet{wu2024moslora} utilizes $A$ and $B$ sub-spaces as the experts but fails to overcome the limitation discussed in the previous section. Another concurrent work, HydraLoRA \citet{tian2024hydralora}, explores an asymmetric design for LoRA. Unlike our work, we consider the interconnection across layers and deploy a model-wide asymmetric design to enable cross-layer connections. This enables the use of adapters of higher rank than a typical LoRA setup while using the same or fewer overall parameters.

\section{Methodology}

\subsection{Downward Projection and Selective Weight Allocation} 

The process is illustrated in the right half of Fig. \ref{fig:cmp-of-lora}. Initially, we use an $A$ to project the input $x \in \mathbb{R}^{N \times C_{\text{in}}}$ into its low-dimensional representation $x' \in \mathbb{R}^{N \times d}$, where $N$ is the sequence length:
\begin{equation}
    x' = xA
\end{equation}
\textbf{To enable more parameter efficiency, the number of $A$s can be set to less than the number of layers in the model by sharing the same $A$ across neighboring layers, as illustrated in Fig. \ref{fig:cmp-of-lora} and discussed in \hyperref[model]{A}}. Inspired by the Mixture of Experts (MoE) paradigm, we employ a router $R \in \mathbb{R}^{N_e \times d}$ to selectively assign weights to all $B$ experts based on their relationship to the current layer's features ($x'$), where $N_e$ represents the number of $B$ experts. A weight set $S \in \mathbb{R}^{N_e}$ is obtained as:
\begin{equation}
    \label{Eqrouter}
    S = \text{softmax}\left(\sum_{i=1}^{N}(x'R^T)_i\right)
\end{equation}
The router selectively mixes the experts based on this data-dependent weight distribution, enabling information integration and expressive adaptation.

\subsection{Weighted Mixture of Experts and Upward Projection} 

Once we obtain the low-dimensional input $x'$, we combine information from all layers using the model-wide shared $B$ experts. One intuitive approach is to feed $x'$ into each $B$ expert and combine their outputs to obtain the additional knowledge $x_{\Delta} \in \mathbb{R}^{N \times C_{\text{out}}}$. However, to address efficiency concerns discussed in Appendix \hyperref[efficiency-discussion]{A.2}, we propose an alternative implementation that is mathematically equivalent but significantly reduces the computational burden, described as follows:

\begin{equation}
    x_{\Delta} = x' \left( \sum_{i=1}^{N_e} S_i \cdot B^i \right)
\label{eq_improve}
\end{equation}

where $S$ is the set of weight scores for the $B$ experts, obtained through selective weight allocation. Since each $S_i$ is a scalar value, the calculation in Eq. \ref{eq_improve} is mathematically equivalent to the intuitive method but with significantly improved efficiency. Therefore, the complete computation flow, with input $x \in \mathbb{R}^{N \times C_{\text{in}}}$ and output $y \in \mathbb{R}^{N \times C_{\text{out}}}$, for an adaptation target module is:

\begin{align}
    y &= xW_0 + s \cdot x_{\Delta}
\end{align}

where $s$ is a scaling factor. By selectively allocating weights and mixing $B$ experts, Lily enables access to all levels of information during adaptation. Each layer's target adaptation modules can consider the status and knowledge from all other layers, resulting in a more expressive and comprehensive adaptation. Meanwhile, thanks to its interconnectivity, Lily can break the low-rank update constraint of LoRA by simply employing a smaller number of adapters with higher ranks.

\section{Experiments}

We validate the effectiveness of Lily across different domains, model sizes (from ViT to LLM), and architectures (Transformers, Mamba), demonstrating its generally strong adaptation capability. Concurrently, we conduct a comprehensive analysis of Lily's intrinsic mechanisms, providing a thorough understanding of how it works. All ranks for Lily are selected from {8, 16, 32}, ensuring that the total parameter count does not exceed that of the baselines. All experiments are conducted on a single RTX 4090 GPU. Additionally, multiple analyses are provided in Appendix \hyperref[sharing-lp-analysis]{C}, \hyperref[finetune-transformer-analysis]{D}, \hyperref[finetune-mamba-analysis]{E}, \hyperref[lr]{F}, \hyperref[off-router]{G}, \hyperref[keep-params-analysis]{H}, \hyperref[diffusion-more]{I}, and \hyperref[attention-map-Lily-lora]{J}.

\subsection{Common Sense Reasoning}

\begin{table*}[t]
\centering
\caption{Commonsense reasoning results for Falcon-Mamba-7B across eight tasks. Bold represents the highest performance for each dataset utilizing PEFT methods. ``$\Delta$'' and ``in'' refer to adaptations of Mamba's \texttt{delta\_proj} and \texttt{in\_proj} parameters, respectively.}
\vspace{-5pt}
\resizebox{1.9\columnwidth}{!}
{
\begin{tabular}{lccccccccccc}
\toprule
\textbf{Model}& \textbf{PEFT} &  \textbf{Params}&\textbf{BoolQ} & \textbf{PIQA} & \textbf{SIQA} & \textbf{HellaSwag} & \textbf{WinoGrande} & \textbf{ARC-e} & \textbf{ARC-c} & \textbf{OBQA} & \textbf{Avg.} \\
\midrule
ChatGPT & -&  -&73.1 & 85.4 & 68.5 & 78.5 & 66.1 & 89.8 & 79.9 & 74.8 & 77.0 \\
\midrule
\multirow{3}{*}{Falcon-Mamba-7B} & LoRA &  3.7M&6.5 & 30.5 & 40.6 & \bf14.9 & 56.4 & 42.2 & 31.8 & 38.4 & 32.7 \\
& Lily ($\Delta \text{ + in}$) &  3.7M&44.9 & \bf66.8 & 65.0 & 10.5 & 57.1 & 78.7 & 64.6 & \bf68.2 & 57.0 \\
& Lily (in) &  \textbf{3.3M}&\bf60.2 & 61.0 & \bf67.3 & 12.9 & \bf61.5 & \bf80.0 & 67.5 & 65.8 & \textbf{59.5} \\
\bottomrule
\end{tabular}
}
\label{tab:commonsense_mamba}
\end{table*}

\begin{table*}[t]
\centering
\caption{Commonsense reasoning results for LLaMA3-8B across eight tasks. $^\dagger$ represents results taken from \citet{liu2024dora} and \citep{wang2024milora}. Bold denotes the highest performance scores for each dataset among different PEFT methods.}
\vspace{-5pt}
\resizebox{1.9\columnwidth}{!}
{
\begin{tabular}{lccccccccccc}
\toprule
\textbf{Model}& \textbf{PEFT} &  \textbf{Params}&\textbf{BoolQ} & \textbf{PIQA} & \textbf{SIQA} & \textbf{HellaSwag} & \textbf{WinoGrande} & \textbf{ARC-e} & \textbf{ARC-c} & \textbf{OBQA} & \textbf{Avg.} \\
\midrule
ChatGPT & -&  -&73.1 & 85.4 & 68.5 & 78.5 & 66.1 & 89.8 & 79.9 & 74.8 & 77.0 \\
\midrule
\multirow{4}{*}{LLaMA3-8B} & LoRA$^\dagger$ &  56M&70.8 & 85.2 & \textbf{79.9} & 91.7 & 84.3 & 84.2 & 71.2 & 79.0 & 80.8 \\
& PiSSA$^\dagger$ &  83.8M&67.1 & 81.1 & 77.2 & 83.6 & 78.9 & 77.7 & 63.2 & 74.6 & 75.4 \\
& MiLoRA$^\dagger$ &  56.6M&68.8 & \bf86.7& 77.2 & \bf92.9 & \bf85.6 & 86.8 & 75.5 & 81.8 & 81.9 \\
& Lily &  \textbf{1.2M}&\bf72.9 & 85.6 & 77.8 & 92.7 & 83.3 & \bf89.7 & \bf77.6 & \bf82.8 & \bf82.8 \\

\bottomrule
\end{tabular}
}
\label{tab:commonsense_llama}
\end{table*}

\textbf{Implementation:} We evaluate Lily on commonsense reasoning with LLMs. For the implementation, we utilize LLaMA3-8B \citep{llama3} and Falcon-Mamba-7B \citep{falconmamba} as backbones. LLaMA3 is a near-SOTA open-source large language model, while Falcon-Mamba is an open-sourced large language model based on the Mamba architecture. Using these models allows us to validate the effectiveness of Lily for fine-tuning LLMs and assess whether this effectiveness can be transferred to architectures beyond Transformers (Mamba, in this case). We fine-tune these models on Commonsense170K \citep{hu-etal-2023-llm} and evaluate the adaptation results on eight multiple-choice problem tasks, including BoolQ \citep{clark-etal-2019-boolq}, PIQA \citep{bisk2020piqa}, SIQA \citep{sap-etal-2019-social}, HellaSwag \citep{zellers2019hellaswag}, WinoGrande \citep{sakaguchi2021winogrande}, ARC-e, ARC-c \citep{clark2018think}, and OBQA \citep{mihaylov2018can}. The compared methods are LoRA for Falcon-Mamba and LoRA \citep{hu2021lora}, PiSSA \citep{meng2024pissa}, and MiLoRA \citep{wang2024milora} for LLaMA3. We only compare LoRA for Falcon-Mamba because tailored PEFT methods for Mamba-based LLMs have not yet been proposed, which is beyond the scope of this paper. Detailed hyper-parameter settings and dataset information are reported in Appendix \hyperref[cmsr-hyper]{B.1.1} and Appendix \hyperref[cmsr-data]{B.2.1}.

\textbf{Results} We report the accuracy in Tables \ref{tab:commonsense_llama} and \ref{tab:commonsense_mamba}. Based on these results, it can be observed that Lily outperforms the other compared PEFT methods with a smaller parameters budget. Specifically, Lily surpasses LoRA by a significant margin on Falcon-Mamba and, on LLaMA3, outperforms both LoRA and MiLoRA. This demonstrates Lily's superior adaptation capability and parameter efficiency in handling commonsense reasoning tasks. Additionally, although performance on Falcon-Mamba is notably lower than that of the baseline and LLaMA3, we believe this discrepancy stems from the inherent limitations of the model rather than any deficiency in Lily, as Lily still significantly outperforms LoRA on Falcon-Mamba while demonstrating robust performance on LLaMA3. These findings also highlight that current state of Mamba-based LLMs generally exhibits inferior performance compared to Transformer-based LLMs such as ChatGPT and LLaMA on many tasks.

\subsection{Natural Language Understanding}

\begin{table*}[t]
\centering
\caption{Various fine-tuning methods applied to RoBERTa Base and RoBERTa Large are evaluated on six datasets from the GLUE benchmark. We present the Matthews correlation coefficient~(MCC) for CoLA, the Pearson correlation coefficient~(PCC) for STS-B, and accuracy~(Acc.) for the remaining tasks. The highest performance for each dataset is highlighted in \textbf{bold}, with all metrics favoring higher values across the six datasets.}
\label{tab:glue}
\vspace{-5pt}
\addtolength{\tabcolsep}{-1pt}
\resizebox{0.8\textwidth}{!}{%
\begin{tabular}{@{}l|c|cccccccc@{}}
\toprule
\textbf{Model \& Method} & \multicolumn{1}{c|}{\begin{tabular}[c]{@{}c@{}}\# Trainable\\ Parameters\end{tabular}} & \begin{tabular}[c]{@{}c@{}}\textbf{SST-2}\\ (Acc.)\end{tabular} & \begin{tabular}[c]{@{}c@{}}\textbf{MRPC}\\ (Acc.)\end{tabular} & \begin{tabular}[c]{@{}c@{}}\textbf{CoLA}\\ (MCC)\end{tabular} & \begin{tabular}[c]{@{}c@{}}\textbf{QNLI}\\ (Acc.)\end{tabular} &  \begin{tabular}[c]{@{}c@{}}\textbf{RTE}\\ (Acc.)\end{tabular} & \begin{tabular}[c]{@{}c@{}}\textbf{STS-B}\\ (PCC)\end{tabular} & \multicolumn{1}{c}{\textbf{Avg.}} \\ \midrule
$\rm{RoB_{base}}$(FFT) & 125M &  94.8 & 90.2 & 63.6 & 92.8  & 78.7 & 91.2 & 85.2 \\
$\rm{RoB_{base}}$(BitFit) & 0.1M  & 93.7 & \textbf{92.7} & 62 & 91.8 & 81.5& 90.8 & 85.4 \\
$\rm{RoB_{base}}$($\text{Adpt}^{\text{D}}$) & 0.3M &  94.2 & 88.5 & 60.8 & 93.1 & 71.5 & 89.7 & 83.0 \\
$\rm{RoB_{base}}$($\text{Adpt}^{\text{D}}$) & 0.9M & 94.7 & 88.4 & 62.6 & 93.0 & 75.9 & 90.3 & 84.2 \\
$\rm{RoB_{base}}$(LoRA) & 0.3M & 94.8 & 89.8 & 63.3 & 92.9 & 78.2 & \textbf{91.5} & 85.1 \\
$\rm{RoB_{base}}$(AdaLoRA) & 0.3M & 94.5 & 88.7 & 62.0 & \bf93.1 & 81.0 & 90.5 & 85.0 \\
$\rm{RoB_{base}}$(DyLoRA) & 0.3M & 94.3 & 89.5 & 61.1 & 92.2 & 78.7 & 91.1 & 84.5 \\
$\rm{RoB_{base}}$(FLoRA) & 0.3M & 91.2 & 85.8 & 65.4 & 92.3 & 65.0 & 87.6 & 81.2 \\
$\rm{RoB_{base}}$(\textbf{Lily}) & 0.3M & \bf95.0 & 90.2 & \textbf{66.0} & 92.5 & \bf81.6 & 90.8 & \bf86.0 \\ \midrule
$\rm{RoB_{large}}$(FF) & 356M &  \bf96.4 & \textbf{90.9} & 68 & 94.7 & 86.6 & \textbf{92.4 }& 88.2 \\
$\rm{RoB_{large}}$($\text{Adpt}^{\text{H}}$) & 0.8M & 96.3 & 87.7 & 66.3 & 94.7 & 72.9 & 91.5 & 84.9 \\
$\rm{RoB_{large}}$(LoRA) & 0.8M & 96.2 & 90.2 & 68.2 & \textbf{94.8} & 85.2 & 92.3 & 87.8 \\
$\rm{RoB_{large}}$(Lily) & 0.5M & 95.6 &\bf90.9 & \bf68.4 & \bf94.8 & \bf88.4 & 91.9 & \bf88.4 \\\bottomrule
\end{tabular}%
}
\vspace{0pt}
\end{table*}

\textbf{Implementation} We evaluate Lily on natural language understanding (NLU) tasks. For the implementation, we use RoBERTa Base \citep{roberta} and RoBERTa Large as the backbones and fine-tune them on tasks from the GLUE benchmark (General Language Understanding Evaluation \citep{glue}), which consists of multiple NLU tasks, including single-sentence classification, similarity and paraphrase, and natural language inference tasks. We compare Lily against several competitive PEFT methods, including BitFit \citep{zaken2021bitfit}, Adapter-Tuning \citep{ada_d,ada_h,ada_l,ada_p}, LoRA \citep{hu2021lora}, DyLoRA \citep{dylora}, FLoRA \citep{hao2024flora}, and AdaLoRA \citep{zhang2023adalora}. Additionally, we utilize full fine-tuning (FFT) as the baseline. Specific hyperparameters and dataset information are provided in Appendix \hyperref[nlu-hyper]{B.1.2} and \hyperlink{nlu-data}{B.2.2}.

\textbf{Results} The results are shown in Table~\ref{tab:glue}. From the table, we can clearly observe that Lily surpasses all the compared PEFT methods by a significant margin, demonstrating its ability to tackle NLU tasks. Among the six tasks, Lily surpasses FFT on four of them when using RoBERTa Base and RoBERTa Large, showcasing its strong approximation ability and high parameter efficiency.

\vspace{0pt}

\begin{figure*}[t]
    \centering
    \includegraphics[width=0.95\linewidth]{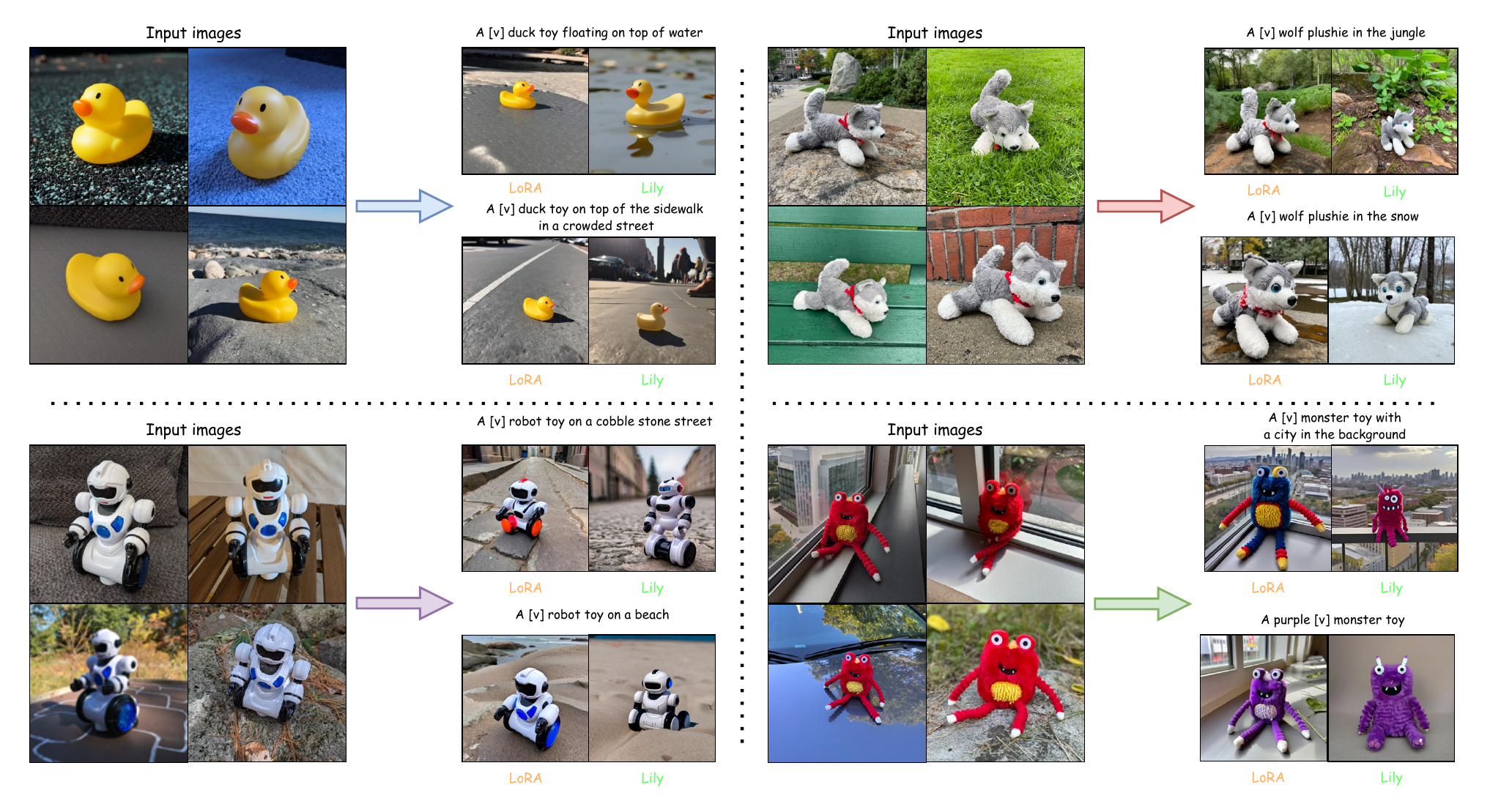}
    \caption{Qualitative results of subject-driven generation. Lily's results align better with prompts, featuring more accurate color, environment, and shape.}
    \label{diff}
\end{figure*}

\subsection{Subject-driven Image Generation}

\textbf{Implementation} We conduct experiments on fine-tuning text-to-image diffusion models for the subject-driven generation task. As the backbone, we use \href{https://huggingface.co/stabilityai/stable-diffusion-xl-base-1.0}{SDXL} and fine-tune it using both LoRA and Lily. First, we fine-tune the model on images paired with text prompts (e.g., “A photo of a [v] duck toy”), each of which includes a unique identifier. Afterward, text prompts containing the identifier are used to generate customized images.

\textbf{Results} The results are presented in Fig. \ref{diff} following the format in \citet{gao2024fourierft} and \citet{wu2024moslora}. From these results, we observe that the images generated by Lily generally align better with the text prompts. For instance, when asked to generate an image of a duck toy floating on water, Lily’s output accurately depicts the designated environment, whereas LoRA’s does not. Additionally, when asked to generate an image of a wolf plushie in the snow, Lily precisely captures the snow around the wolf, while LoRA fails to do so. These observations demonstrate Lily’s excellent performance in text-to-image generation with more expressive adaptation. Additional generated results are provided in Appendix \hyperref[diffusion-more]{I}.

\subsection{Visual Adaptation Benchmark}

\begin{table*}[t]
\caption{Full results of Lily on ViT-B pre-trained on ImageNet-21K for the VTAB-1K benchmark, with averages computed based on group-wise results. \textbf{Bold} indicates the best performance.}
\vspace{-10pt}
\begin{center}
\setlength{\tabcolsep}{0.3pt}
\scalebox{0.75}{
\begin{tabular}{
p{2.30cm}<{}
p{1.25cm}<{\centering}
p{0.75cm}<{\centering}
|
p{0.75cm}<{\centering}
p{0.75cm}<{\centering}
p{0.75cm}<{\centering}
p{0.75cm}<{\centering}
p{0.75cm}<{\centering}
p{0.75cm}<{\centering}
p{0.75cm}<{\centering}
|
p{0.75cm}<{\centering}
p{0.75cm}<{\centering}
p{0.75cm}<{\centering}
p{0.75cm}<{\centering}
|
p{0.75cm}<{\centering}
p{0.75cm}<{\centering}
p{0.75cm}<{\centering}
p{0.75cm}<{\centering}
p{0.75cm}<{\centering}
p{0.75cm}<{\centering}
p{0.75cm}<{\centering}
p{0.75cm}<{\centering}
}
\toprule
\multicolumn{3}{c|}{}&\multicolumn{7}{c|}{\textbf{Natural}}&\multicolumn{4}{c|}{\textbf{Specialized}}&\multicolumn{8}{c}{\textbf{Structured}}\\
&\multicolumn{1}{c}{\STAB{\rotatebox[origin=c]{90}{Params(M)}}}
&\multicolumn{1}{c|}{\STAB{\rotatebox[origin=c]{90}{Average}}}
&\multicolumn{1}{c}{\STAB{\rotatebox[origin=c]{90}{\textbf{Cifar100}}}}
&\multicolumn{1}{c}{\STAB{\rotatebox[origin=c]{90}{\textbf{Caltech101}}}}
&\multicolumn{1}{c}{\STAB{\rotatebox[origin=c]{90}{\textbf{DTD}}}}
&\multicolumn{1}{c}{\STAB{\rotatebox[origin=c]{90}{\textbf{Flowers102}}}}
&\multicolumn{1}{c}{\STAB{\rotatebox[origin=c]{90}{\textbf{Pets}}}}
&\multicolumn{1}{c}{\STAB{\rotatebox[origin=c]{90}{\textbf{SVHN}}}}
&\multicolumn{1}{c|}{\STAB{\rotatebox[origin=c]{90}{\textbf{Sun397}}}}
&\multicolumn{1}{c}{\STAB{\rotatebox[origin=c]{90}{\textbf{Camelyon}}}}
&\multicolumn{1}{c}{\STAB{\rotatebox[origin=c]{90}{\textbf{EuroSAT}}}}
&\multicolumn{1}{c}{\STAB{\rotatebox[origin=c]{90}{\textbf{Resisc45}}}}
&\multicolumn{1}{c|}{\STAB{\rotatebox[origin=c]{90}{\textbf{Retinopathy}}}}
&\multicolumn{1}{c}{\STAB{\rotatebox[origin=c]{90}{\textbf{Clevr-Count}}}}
&\multicolumn{1}{c}{\STAB{\rotatebox[origin=c]{90}{\textbf{Clevr-Dist}}}}
&\multicolumn{1}{c}{\STAB{\rotatebox[origin=c]{90}{\textbf{DMLab}}}}
&\multicolumn{1}{c}{\STAB{\rotatebox[origin=c]{90}{\textbf{KITTI-Dist}}}}
&\multicolumn{1}{c}{\STAB{\rotatebox[origin=c]{90}{\textbf{dSpr-Loc}}}}
&\multicolumn{1}{c}{\STAB{\rotatebox[origin=c]{90}{\textbf{dSpr-Ori}}}}
&\multicolumn{1}{c}{\STAB{\rotatebox[origin=c]{90}{\textbf{sNORB-Azim}}}}
&\multicolumn{1}{c}{\STAB{\rotatebox[origin=c]{90}{\textbf{sNORB-Ele}}}}\\
\specialrule{0em}{1pt}{1pt}
\hline
\specialrule{0em}{1pt}{1pt}
\multicolumn{22}{l}{\emph{Conventional Fine-Tuning}}\\
\hline
\specialrule{0em}{1pt}{1pt}
FFT&86&68.9&68.9&87.7&64.3&97.2&86.9&87.4&38.8&79.7&95.7&84.2&73.9&56.3&58.6&41.7&65.5&57.5&46.7&25.7&29.1 \\
LP&0&57.6&64.4&85.0&63.2&97.0&86.3&36.6&51.0&78.5&87.5&68.5&74.0&34.3&30.6&33.2&55.4&12.5&20.0&9.6&19.2\\
\hline
\specialrule{0em}{1pt}{1pt}
\multicolumn{22}{l}{\emph{PEFT methods}}\\
\hline
\specialrule{0em}{1pt}{1pt}
AdaptFormer&0.588&76.8&74.0&92.2&71.7&\bf99.3&\bf91.7&88.9&56.4&87.2&95.1&\bf85.7&\bf75.9&\bf84.2&62.2&53.0&81.0&87.1&53.6&35.3&42.3 \\
Bi-LoRA&1.180&76.7&72.1&91.7&71.2&99.1&91.4&\bf90.2&55.8&87.0&\bf95.4&85.5&75.5&83.1&64.1&52.2&81.3&86.4&53.5&36.7&44.4\\
LoRA&1.180&76.4&72.5&91.5&71.9&99.1&91.4&89.6&56.0&87.6&95.3&84.0&75.0&83.6&64.3&51.6&80.9&86.0&51.8&\bf36.8&42.3\\
FourierFT&0.936&72.7&69.1&88.8&71.9&99.0&91.0&79.0&55.6&84.9&93.0&83.2&74.9&70.7&61.1&45.2&74.8&78.0&53.0&24.8&30.8 \\
MoRA&1.058&75.4&72.1&90.0&71.7&99.2&91.1&90.1&56.0&87.1&94.8&85.1&75.4&76.7&62.3&49.7&78.3&83.1&53.0&34.5&34.5 \\
Lily&0.318&\bf77.3&\bf73.9&\bf93.0&\bf72.9&\bf99.3&91.6&89.0&\bf56.6&\bf87.9&95.2&84.9&75.7&83.9&\bf65.4&\bf53.4&\bf81.6&\bf88.2&\bf54.5&\bf37.0&\bf45.4\\
\bottomrule
\end{tabular}
}
\end{center}
\label{tab:vtab-vit}
\end{table*}
\vspace{-5pt}
\begin{table*}[ht]
\caption{Full results of Lily on Vim-S pre-trained on ImageNet-1K for the VTAB-1K benchmark, with averages calculated within each group. * denotes linear probing results from \citet{tu2023VQT}. For fair comparison, we also use ViT-B pre-trained on ImageNet-1K. \textbf{Bold} indicates best performance among Vim-based PEFT methods.}
\vspace{-10pt}
\begin{center}
\setlength{\tabcolsep}{0.3pt}
\scalebox{0.75}{
\begin{tabular}{
p{2.30cm}<{}
p{1.25cm}<{\centering}
p{0.75cm}<{\centering}
|
p{0.75cm}<{\centering}
p{0.75cm}<{\centering}
p{0.75cm}<{\centering}
p{0.75cm}<{\centering}
p{0.75cm}<{\centering}
p{0.75cm}<{\centering}
p{0.75cm}<{\centering}
|
p{0.75cm}<{\centering}
p{0.75cm}<{\centering}
p{0.75cm}<{\centering}
p{0.75cm}<{\centering}
|
p{0.75cm}<{\centering}
p{0.75cm}<{\centering}
p{0.75cm}<{\centering}
p{0.75cm}<{\centering}
p{0.75cm}<{\centering}
p{0.75cm}<{\centering}
p{0.75cm}<{\centering}
p{0.75cm}<{\centering}
}
\toprule
\multicolumn{3}{c|}{}&\multicolumn{7}{c|}{\textbf{Natural}}&\multicolumn{4}{c|}{\textbf{Specialized}}&\multicolumn{8}{c}{\textbf{Structured}}\\
&\multicolumn{1}{c}{\STAB{\rotatebox[origin=c]{90}{Params(M)}}}
&\multicolumn{1}{c|}{\STAB{\rotatebox[origin=c]{90}{Average}}}
&\multicolumn{1}{c}{\STAB{\rotatebox[origin=c]{90}{\textbf{Cifar100}}}}
&\multicolumn{1}{c}{\STAB{\rotatebox[origin=c]{90}{\textbf{Caltech101}}}}
&\multicolumn{1}{c}{\STAB{\rotatebox[origin=c]{90}{\textbf{DTD}}}}
&\multicolumn{1}{c}{\STAB{\rotatebox[origin=c]{90}{\textbf{Flowers102}}}}
&\multicolumn{1}{c}{\STAB{\rotatebox[origin=c]{90}{\textbf{Pets}}}}
&\multicolumn{1}{c}{\STAB{\rotatebox[origin=c]{90}{\textbf{SVHN}}}}
&\multicolumn{1}{c|}{\STAB{\rotatebox[origin=c]{90}{\textbf{Sun397}}}}
&\multicolumn{1}{c}{\STAB{\rotatebox[origin=c]{90}{\textbf{Camelyon}}}}
&\multicolumn{1}{c}{\STAB{\rotatebox[origin=c]{90}{\textbf{EuroSAT}}}}
&\multicolumn{1}{c}{\STAB{\rotatebox[origin=c]{90}{\textbf{Resisc45}}}}
&\multicolumn{1}{c|}{\STAB{\rotatebox[origin=c]{90}{\textbf{Retinopathy}}}}
&\multicolumn{1}{c}{\STAB{\rotatebox[origin=c]{90}{\textbf{Clevr-Count}}}}
&\multicolumn{1}{c}{\STAB{\rotatebox[origin=c]{90}{\textbf{Clevr-Dist}}}}
&\multicolumn{1}{c}{\STAB{\rotatebox[origin=c]{90}{\textbf{DMLab}}}}
&\multicolumn{1}{c}{\STAB{\rotatebox[origin=c]{90}{\textbf{KITTI-Dist}}}}
&\multicolumn{1}{c}{\STAB{\rotatebox[origin=c]{90}{\textbf{dSpr-Loc}}}}
&\multicolumn{1}{c}{\STAB{\rotatebox[origin=c]{90}{\textbf{dSpr-Ori}}}}
&\multicolumn{1}{c}{\STAB{\rotatebox[origin=c]{90}{\textbf{sNORB-Azim}}}}
&\multicolumn{1}{c}{\STAB{\rotatebox[origin=c]{90}{\textbf{sNORB-Ele}}}}\\
\specialrule{0em}{1pt}{1pt}
\hline
\specialrule{0em}{1pt}{1pt}
\multicolumn{22}{l}{\emph{Conventional Fine-Tuning}}\\
\hline
\specialrule{0em}{1pt}{1pt}
FFT-Vim&26&70.1&47.7&89.4&64.2&89.0&87.7&90.6&35.1&84.5&93.9&81.0&74.5&67.5&52.9&47.3&78.9&75.3&53.9&33.3&29.4 \\
FFT-ViT&86&69.9&49.4&89.3&65.5&91.7&89.1&91.4&33.5&85.9&93.6&85.4&74.3&54.7&55.2&48.7&79.7&68.2&49.7&31.5&27.7\\
LP-Vim&0&55.3&40.9&83.3&57.3&66.3&86.3&38.4&34.6&79.0&87.6&65.0&73.6&36.3&35.1&33.3&64.8&23.0&21.6&15.1&21.7\\
LP-ViT&0&66.4&50.6&85.6&61.4&79.5&86.5&40.8&38.0&79.7&91.5&71.7&65.5&41.4&34.4&34.1&55.4&18.1&26.4&16.5&24.8\\
\hline
\specialrule{0em}{1pt}{1pt}
\multicolumn{22}{l}{\emph{PEFT on ViT}}\\
\hline
\specialrule{0em}{1pt}{1pt}
AdaptFormer&0.147&72.4&56.2&89.6&67.2&91.2&91.1&85.9&42.1&85.4&94.6&84.0&74.3&75.8&58.6&48.6&79.6&81.6&53.7&29.6&35.2\\
LoRA&0.295&72.5&56.4&89.0&66.9&91.2&90.4&86.9&41.5&85.4&95.1&84.1&75.2&75.8&61.7&47.7&80.5&80.4&52.0&29.4&35.7\\
\hline
\specialrule{0em}{1pt}{1pt}
\multicolumn{22}{l}{\emph{PEFT on Vim}}\\
\hline
\specialrule{0em}{1pt}{1pt}
LoRA&0.054&70.1&57.5&87.7&64.4&86.0&90.0&85.7&39.8&82.2&93.8&79.6&72.5&78.6&56.5&42.0&80.5&71.8&51.0&28.4&32.6\\
Lily-S&0.074&71.4&\bf58.2&88.5&65.6&87.1&\bf90.7&87.5&40.4&83.3&94.1&79.7&73.8&81.2&57.3&44.1&80.9&79.3&54.1&30.0&33.7 \\ 
Lily-L&0.196&\bf72.3&57.8&\bf89.4&\bf66.2&\bf87.8&90.5&\bf88.1&\bf40.5&\bf84.1&\bf94.3&\bf81.3&\bf75.1&\bf81.6&\bf57.8&\bf46.5&\bf81.0&\bf82.9&\bf55.2&\bf32.1&\bf34.8 \\ 
\bottomrule
\end{tabular}
}
\end{center}
\label{tab:vtab-vim}
\end{table*}

\textbf{Implementation} We assess Lily on the Visual Task Adaptation Benchmark (VTAB-1K \cite{zhai2019vtab}), a suite of 19 visual tasks spanning diverse domains and semantics, to test its general visual adaptation capability. Tasks are categorized into Natural, Specialized, and Structured, and are all formulated as classification problems for consistent model evaluation. We conduct two sets of experiments: one focusing on adaptation effectiveness on the Vision Transformer (ViT \citep{dosovitskiy2020ViT}) and the other on Vision Mamba (Vim \citep{zhu2024visionMamba}), demonstrating Lily's architecture-agnostic capabilities. For ViT, we use ViT-B pre-trained on ImageNet-21K \citep{deng2009imagenet}, and for Vim, we use Vim-s pre-trained on ImageNet-1K. To fairly compare ViT and Vim architectures, we implement LoRA \citep{hu2021lora} and AdaptFormer \citep{chen2022adaptformer} on ViT-B pre-trained on ImageNet-1K. In the ViT experiments, we compare Lily with LoRA, AdaptFormer, FourierFT \citep{gao2024fourierft}, and MoRA \citep{jiang2024mora}; in the Vim experiments, we focus on contrasting architectural differences and, therefore, use only LoRA as the baseline. All experiments include full fine-tuning (FFT) and linear probing as baselines. For Vim, we implement two versions of Lily: Lily-S (Small) and Lily-L (Large), with different hyperparameter settings to either reduce the parameter count (Lily-S) or maximize performance (Lily-L). For Lily on ViT, the reported results are obtained from adapting both the self-attention and the M$A$ module in the Transformer. Regarding the performance of the fine-tuned module, we conduct additional experiments in Appendix \hyperref[finetune-transformer-analysis]{D}. Detailed experimental settings and dataset information are provided in Appendix \hyperref[cv-hyper]{B.1.3} and \hyperref[cv-data]{B.2.3}.

\textbf{Results} The results are shown in Tables \ref{tab:vtab-vit} and \ref{tab:vtab-vim}. For ViT, Lily significantly outperforms all compared PEFT methods while also offering improved parameter efficiency. In contrast, the performance on the Vim backbone is generally lower than that on ViT; for instance, LoRA on ViT performs better than LoRA on Vim. We argue that this difference is due to variations in architecture design and overall model size. However, Lily's strong adaptation performance allows it to match or exceed the performance of other PEFT methods on ViT and to significantly outperform LoRA on Vim (with both Lily-S and Lily-L surpassing LoRA by a significant margin). This demonstrates Lily's architecture-agnostic capability, highlighting its potential across various model architectures. Overall, Lily achieves excellent visual adaptation performance while maintaining architecture-agnosticity and high parameter efficiency.

\subsection{Understanding Lily}

\subsubsection{Does It Have High-Rank Weight Updates?}

\begin{figure}[t]
    \centering
    \includegraphics[width=0.99\linewidth]{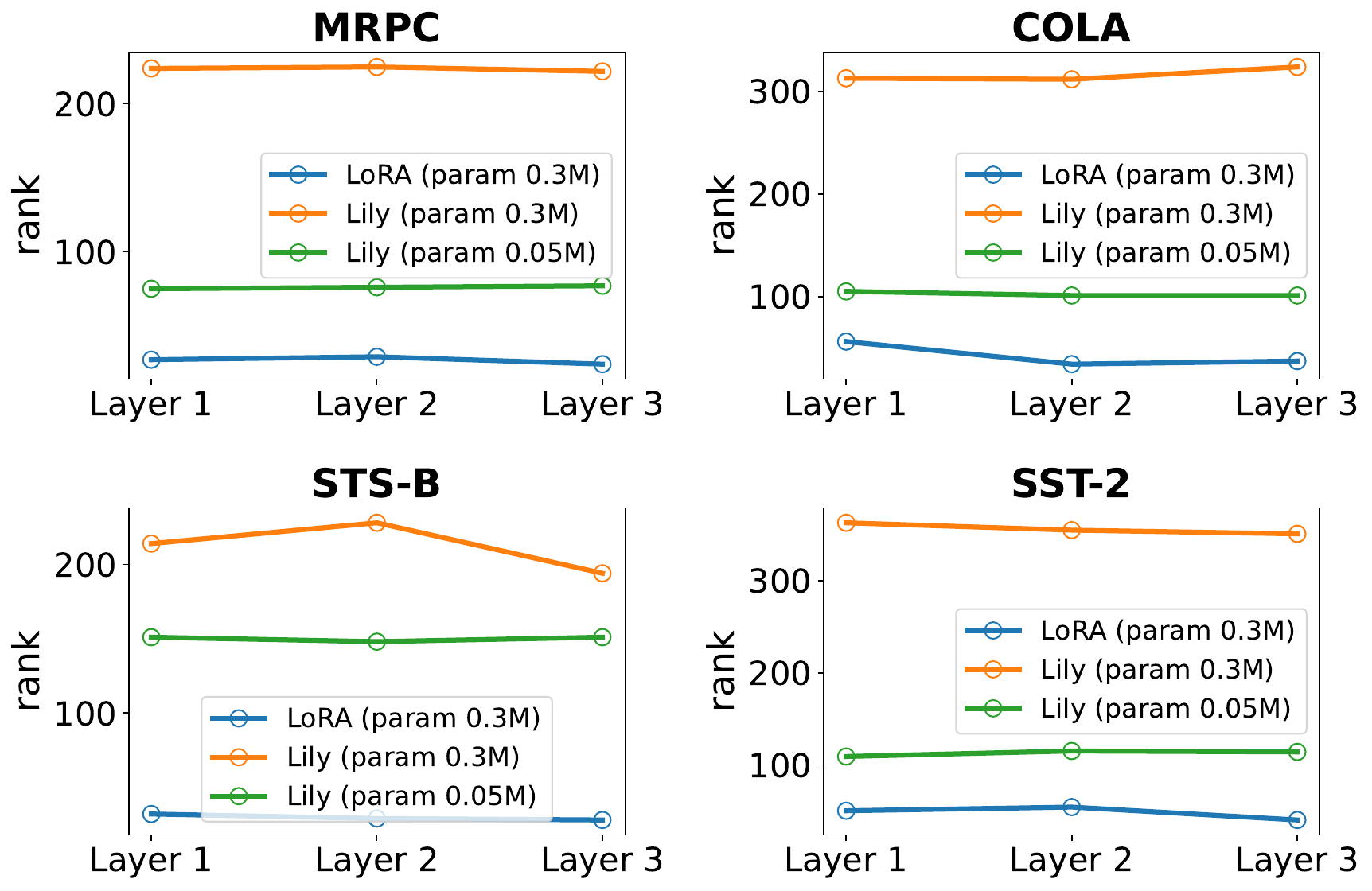}
    \caption{Actual rank of the weight updates. The weight updates are of shape \( 768 \times 768 \). We run 20 epochs for COLA, MRPC, and STS-B, and 3 epochs for SST-2. It can be easily observed that the weight updates from Lily have notably higher rank than those from LoRA. Note that the reported rank is computed from accumulated weight updates over multiple epochs.}
    \label{fig:rk}
\end{figure}

The interconnected and asymmetric structure of Lily enables a flexible allocation of the parameter budget, thereby allowing weight updates with higher ranks across all layers. To validate this claim, we provide an empirical analysis, as shown in Fig. \ref{fig:rk}. Specifically, we run four tasks from the NLU experiment and measure the rank of the weight updates for the query transformation matrix $W_q$ in the first three layers. We use a small number of matrices $A$ and $B$ (2 or 3) with a rank of 32 to match the parameter count of LoRA, which uses adapters with a rank set to 8. Specific hyperparameter settings can be found in Appendix \hyperref[nlu-hyper]{B.1.2}.

From the results, we observe that the rank of the weight updates from Lily is generally notably larger than that of LoRA when using a similar number of parameters. Meanwhile, the weight updates from Lily still exhibit a higher rank compared to those of LoRA even when using only $16.7\%$ of LoRA's parameters. This empirical analysis essentially validates our claim that Lily achieves high-rank updates with the same parameter budget. We attribute this to the model-wide sharing and the cross-layer asymmetric design, which facilitate a flexible allocation of the parameter budget.

\begin{figure*}[ht]
    \centering
    \includegraphics[width=0.95\linewidth]{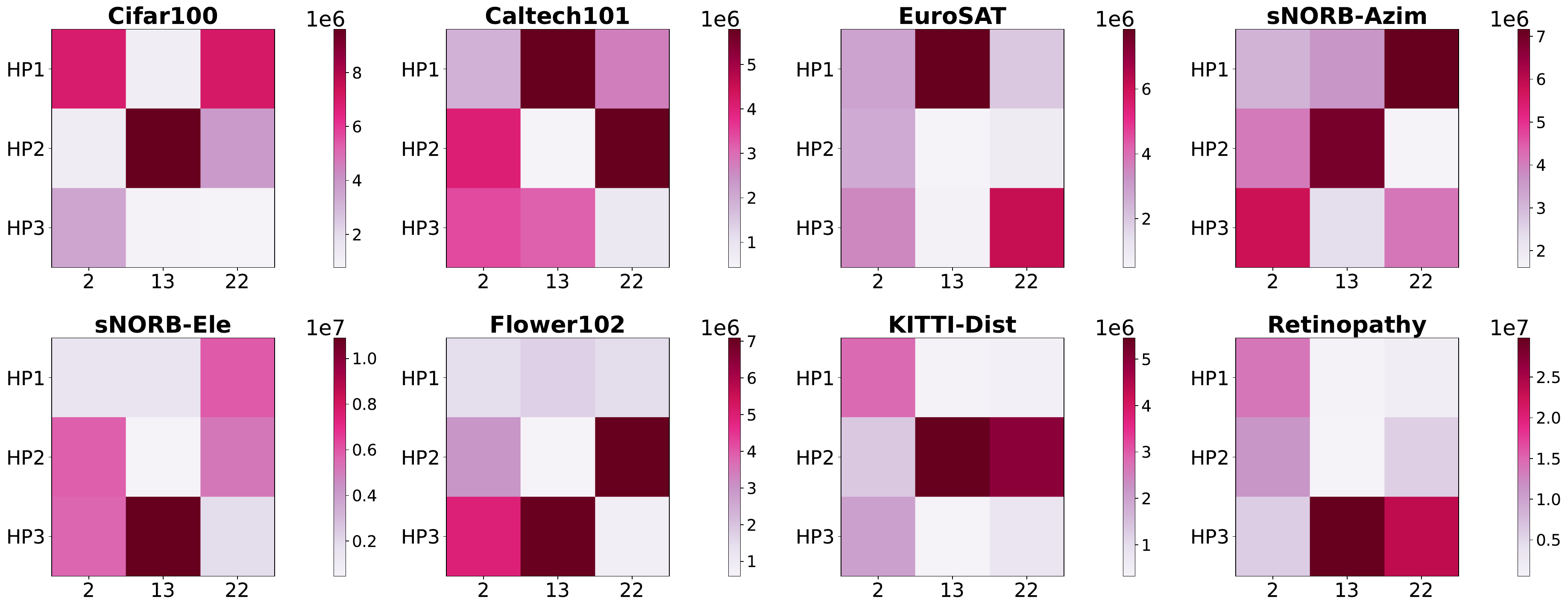}
    \caption{Visualization of accumulated assigned weight for $B$ experts by a router across various layers. Example here uses layer of index 2, 13 and 22 to represent shallow, middle and deep layers. The reported values are based on the accumulated router outputs over multiple epochs.}
    \label{fig:heatmap}
\end{figure*}

\begin{figure*}[t]
    \centering
    \includegraphics[width=0.99\linewidth]{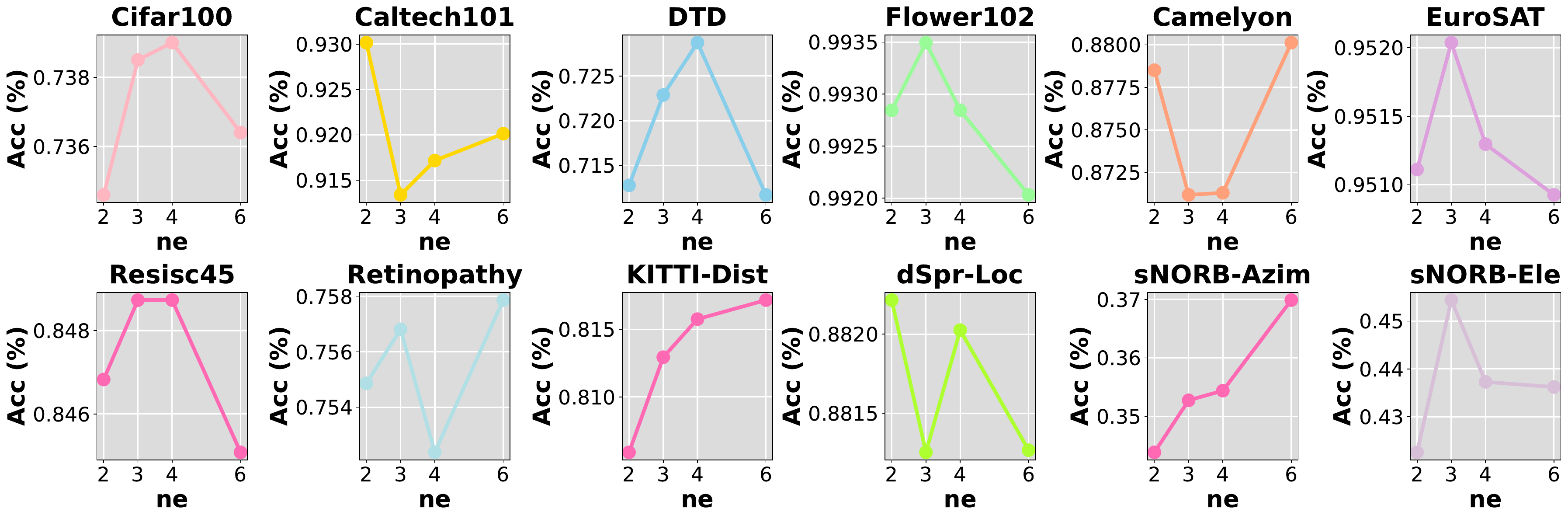}
    \caption{Impact of attention granularity (i.e., the choice of how many $A$s and $B$s) on the performance. We choose 12 out of 19 tasks from VTAB-1K for a comprehensive understanding.}
    \label{fig:attention_gran}
\end{figure*}

\begin{figure}[ht]
    \centering
    \includegraphics[width=0.99\linewidth]{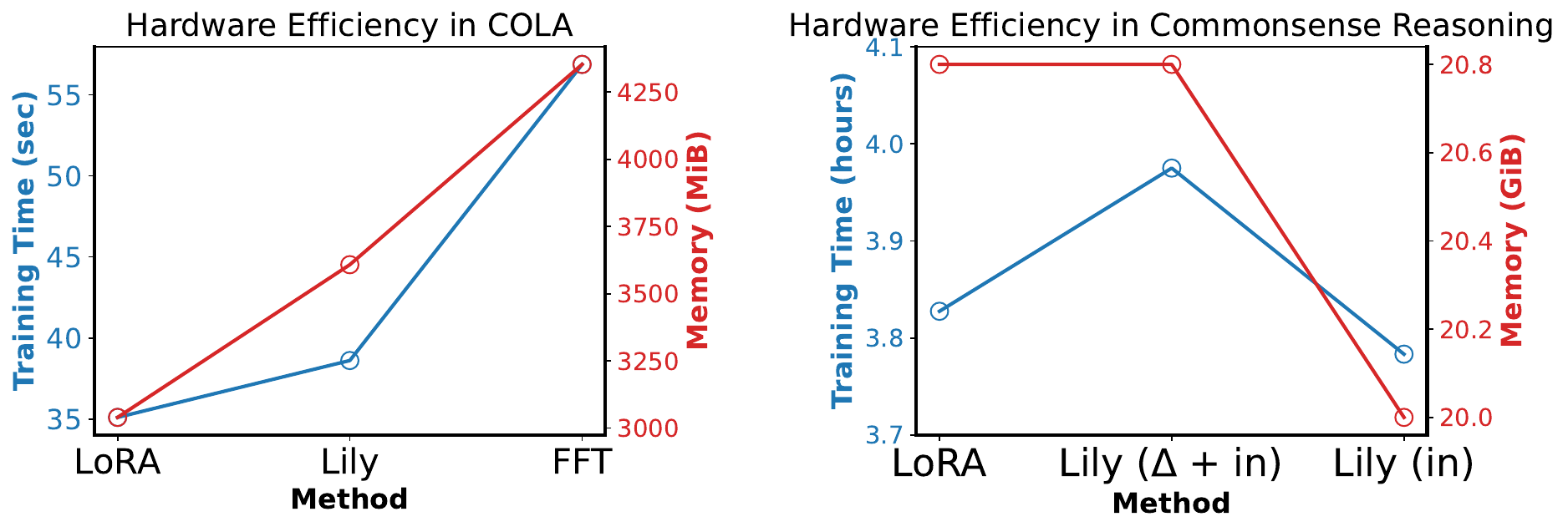}
    \caption{Hardware efficiency of Lily compared to LoRA. We run 10 epochs for COLA. We report the training time and memory consumption. It can be observed that Lily generally performs on par with LoRA in terms of hardware efficiency.}
    \label{fig:efficiency}
    \vspace{-15pt}
\end{figure}

\subsubsection{What's the Influence of Adapter Granularity?}

The number of experts in the model-wide $B$ module can be freely set, and the number of $A$s can also be flexibly determined by sharing across the same level of layers, as introduced in Appendix \hyperref[model]{A.1}. Therefore, we analyze the impact of these choices on performance. We denote the number of $A$ experts and $B$ experts as \textit{ne}\_1 and \textit{ne}\_2, respectively. For simplicity, we set them equal in the experiments and denote this common value as \textbf{\textit{ne}}. We refer to the number of layers each expert attends to as adapter granularity. As the value of \textit{ne} increases, the adapter granularity becomes finer. As shown in Fig. \ref{fig:attention_gran}, the results from the VTAB-1K benchmark indicate different patterns. For instance, on the DTD dataset, the best performance is achieved when \textit{ne} is 4, while on sNORB-Azim, performance increases as \textit{ne} increases. Increasing \textit{ne} leads to more parameters and finer adapter granularity; however, finer adapter granularity does not necessarily translate to better overall performance. For example, on Resisc45, DTD, Cifar100, sNORB-Ele, dsPr-LoC, Flowers102, and EuroSAT, the negative impact of increasingly finer adapter granularity eventually outweighs the benefits of the additional parameters, leading to a decrease in overall performance. In other tasks, different patterns may occur because the positive effect of adapter granularity on performance is consistently strong, or its negative effect is insufficient to offset the benefits of increased parameters, resulting in generally improved performance with higher \textit{ne}. This phenomenon provides an important insight: for most tasks, simply increasing the number of parameters may not lead to better performance. Instead, only when adapter granularity and the number of parameters reach an optimal trade-off can we achieve the best performance.

\subsubsection{Does It Exhibit Selectivity?}

Lily uses routers to assign varying weights to different $B$ experts, thereby achieving selective information combination. We illustrate this selectivity in Fig. \ref{fig:heatmap}. We use a setup with three $B$ experts and select three layer levels (1, 13, 22) to calculate the total weight assigned to each expert. The results reveal clear selectivity: for different layers, the router assigns significantly different weights to the $B$ experts. For instance, on Cifar100, the middle layer is predominantly dominated by $B$ 2, whereas the deep layer is primarily dominated by $B$ 1 and $B$ 2. In contrast, on Retinopathy, both the middle and deep layers are dominated by $B$ 3. This selectivity ensures that, even when different layers share information, the inherent differences between layers are still taken into account, making the adaptation more flexible and comprehensive.

\subsubsection{What's the Hardware Efficiency?}

The dynamics of Lily obviously introduce complexity to the design of LoRA. In this section, we analyze how this affects the hardware efficiency of Lily compared to LoRA. We use the COLA task from the NLU experiments with RoBERTa-Base and run for 10 epochs. Additionally, we also report the runtime and GPU memory consumption in the Falcon-Mamba experiment.

The results are shown in Fig. \ref{fig:efficiency}, from which we can observe that the hardware efficiency of Lily is comparable to LoRA. Specifically, Lily slightly underperforms LoRA in the NLU experiment but performs on par with LoRA in the LLM experiment. In general, the introduced complexity of Lily does not prevent it from being an more effective PEFT method that is also hardware-friendly.

\section{Conclusion}

In this paper, we propose Low-Rank Interconnected Adaptation (Lily), a novel framework for efficient fine-tuning via the interconnectivity of adapters. Lily enables each layer to access information from others during adaptation through a hierarchical structure. Additionally, it successfully overcomes the low-rank update limitation of LoRA, enabling high-rank updates and, therefore, better adaptation capability under the same parameter budget. Our approach consistently improves performance across various modalities, model sizes, and architectures, surpassing existing methods while maintaining enhanced efficiency. In summary, Lily's versatility and efficiency make it a promising approach for a wide range of applications.

\section{Limitations}

Although Lily has been experimentally evaluated in a wide range of scenarios, we have not explored all possible applications where PEFT could be used. These potential areas are left as directions for future work.  

\section{Ethics Statement}

This work is an improvement upon LoRA. However, it could potentially be used for fine-tuning diffusion models or large language models (LLMs) for generating malicious content.

\bibliography{custom}
\clearpage
\appendix

\section*{Appendix}
\section{More discussion about Lily}
\subsection{Model Structure and Design Intuition of Lily}
\label{model}
Within the overall framework of Lily, we delve into specific implementation details and model design insights. First, we establish the relationship between $A$ and $B$: $A$ is confined to specific levels of layers, capturing features that enable the router to selectively assign weights to the $B$ experts. In contrast, $B$ is a model-wide module comprising multiple experts, each of which contains information from a particular level of layers. 

We highlight several key aspects that are not heavily discussed in the methodology section:

\subsubsection{Number of $A$s} 

Since $A$ is limited to specific layers, the simplest approach would be to place an $A$ at each layer of the module to be adapted (e.g., the query transformation in MHSA). However, this setup may not be necessary, as the importance of each layer varies, and many layers have significantly lower importance than others \citep{zhang2023adalora}. 

To achieve greater parameter efficiency, we can use fewer $A$s, with each $A$ focusing on a level of layers rather than a single layer. For example, an $A$ can focus on shallow layers (e.g., layers 0, 1, 2, etc.) or deep layers. To enable a single $A$ to handle multiple layers, we can share an $A$ across multiple layers. By doing so, we eliminate the redundancy of placing an $A$ at each layer, reduce the number of parameters, and improve efficiency. 

\textbf{This is exactly the strategy adopted in most of the experiments.}

\subsubsection{Number of $B$ Experts} 

Regarding $B$, the number of experts can be set arbitrarily, allowing for more flexible configurations. In our experiments, for the sake of simplicity, we set the number of $B$ experts equal to the number of $A$s, thereby equating the granularity of $A$ and $B$.

\subsubsection{Routers Setup} 

There are multiple possible configurations for the router. First, we can bind the router to $B$, resulting in only one router per model. However, since the number of parameters in the router is relatively small, having only one router per model may not provide significant selectivity. Therefore, we can also bind the router to $A$, configuring a separate router for each $A$. 

Most of our experiments use the latter setup. However, in the vision experiments on Vim, we adopt the single-router and no-lp-sharing setup to evaluate its effectiveness. The results indicate that this setup also performs well. 

As future work, we can further verify the effectiveness of the latter setup on Vim, which may potentially lead to superior performance.

\subsubsection{Hyperparameters}  

We detail the hyperparameters used in Lily. Specifically, we use \textbf{Lily\_r} to represent the hidden dimension of the projectors: $A$s and $B$s. It serves the same function as $r$ in LoRA. We use \textbf{Lily\_s} to represent the scaling factor used by Lily, which is primarily searched within the range \{0.01, 0.1, 1.0, 10.0, 100.0\}.  

We use \textbf{\textit{ne}\_1} to denote the number of $A$s used in the model. Since $A$s can be shared, as discussed in the previous section, \textit{ne}\_1 does not necessarily equal the number of layers in the model. Similarly, we use \textbf{\textit{ne}\_2} to represent the number of $B$ experts in the model-wide $B$ module.  

In our experiments, we set \textit{ne}\_1 = \textit{ne}\_2 to enhance parameter efficiency and maintain simplicity.

\subsubsection{Design Intuition}  

Lily employs a hierarchical structure to enable updates with higher ranks than LoRA. However, simply connecting all $B$s equally to $A$s does not yield the best performance. From the perspective of feature and information utilization across layers, merely aggregating all $B$s for an $A$ ignores the distinctiveness of features from the current layers. Meanwhile, this approach reduces the variability in the combinations of gradient projection matrices (since $S_i$ and $C_{i,j}$ become constants), making the rank of the weight update higher than that of LoRA (as multiple distinct random matrices are used), but still not high enough for optimal performance due to the lack of variability in the combination process.  

To address this, we introduce selectivity into the interconnectivity, as discussed below, making the combination of $B$s data-dependent. This ensures that each $S_i$ is unique across time steps, enabling updates with even higher ranks. This approach is similar to that of \citet{hao2024flora}, where a random matrix is constantly resampled to maintain high-rank updates. We further analyze this in Appendix \hyperref[off-router]{G}.

\subsection{Efficient Implementation for Weighted Combination}
\label{efficiency-discussion}

A straightforward implementation of the weighted combination in Lily is to pass the inputs through all the experts and then sum the results. This approach requires $N_e$ matrix multiplications, $N_e$ scalar multiplications, and $N_e$ matrix additions. Despite its intuitive nature, the computational burden of this method is quite substantial.  

However, Eq. \ref{eq_improve}, which is adopted in Lily, requires only $N_e$ scalar multiplications, $N_e$ matrix additions, and a single matrix multiplication. This optimization eliminates approximately $N_e$ matrix multiplications, which can significantly reduce computational costs as the model size and the number of adaptation targets increase.  

For an input $x'$ of size $\mathbb{R}^{N \times d}$ and a projection matrix $P_H \in \mathbb{R}^{d \times C}$, the floating-point operations (FLOPs) of these two implementations are:  

\begin{equation}
    \begin{aligned}
         \text{FLOPs} &=  \sum_{i=1}^{N_e} (2NdC) + \sum_{i=1}^{N_e} (dC) + \sum_{i=1}^{N_e} (NC)  \\
                      &= N_e \times (2NdC + dC + NC), &&\\
         \text{FLOPs} &= 2\sum_{i=1}^{N_e} (dC) + 2NdC \\
                      &= 2dC \times (N + N_e), && 
    \end{aligned}
\end{equation}

From this, we can easily observe that the approach adopted by Lily requires fewer computations, thereby improving both speed and efficiency during the fine-tuning process. Under the setting of $N=1024, d=16, C=768, N_e=4$, the FLOPs for the intuitive approach amount to $0.104$ GFLOPs, whereas for Lily, it is merely $0.025$ GFLOPs, potentially leading to a $4\times$ speedup.

\subsection{Actual Implementation of Lily}
\label{actual-code}

\begin{figure*}[t]
    \centering
    \includegraphics[width=\linewidth]{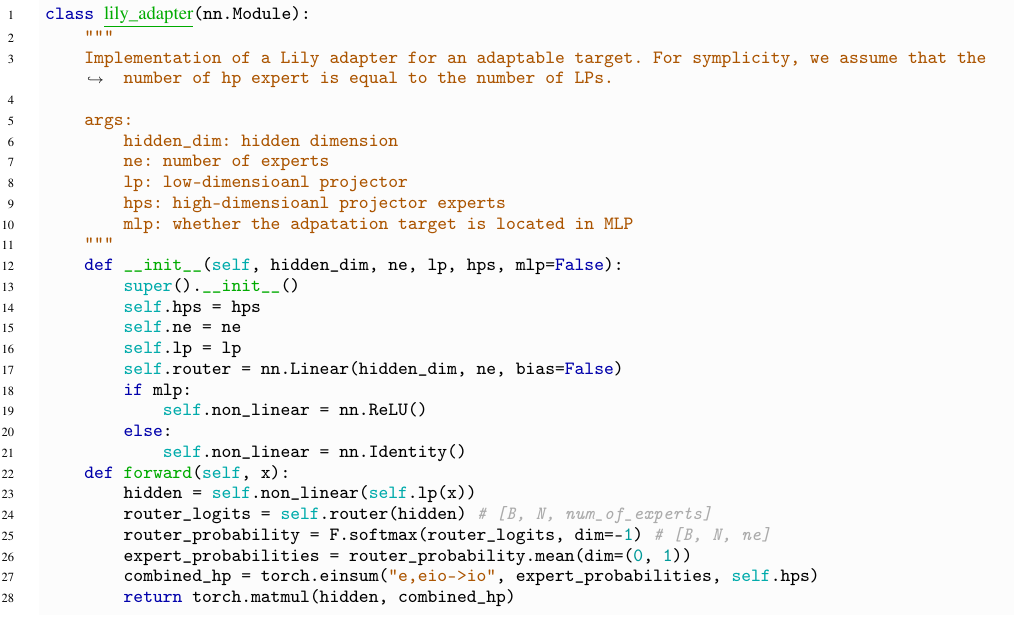}
    \caption{Implementation of Lily in the VTAB-1K benchmark.}
    \label{actual-code-alg}
\end{figure*}

We present the actual implementation of Lily in Fig. \ref{actual-code-alg}. In this example, we showcase its implementation for visual adaptation tasks, specifically in the VTAB-1K benchmark. For large language models (LLMs), the implementation is slightly more complex due to modifications to the Hugging Face PEFT library \citep{peft}, but the fundamental adaptation process remains the same.  

Specifically, given an input, we first use the corresponding $A$ of the current layer to project it into a low-dimensional representation. This low-dimensional representation is then used to selectively assign weights to the $B$ experts. Once the weights for all experts are obtained, we proceed to combine these $B$ experts accordingly, as discussed in Appendix \hyperref[efficiency-discussion]{A.2}. After the weighted combination, the combined $B$ is used to project the low-dimensional representation back into a high-dimensional space, thereby incorporating the additional knowledge gained through adaptation.

\section{Experimental Settings}
\subsection{Hyperparameters}

A detailed description of the hyperparameters used in Lily is provided in Appendix \hyperref[model]{A.1}.  

\subsubsection{Commonsense Reasoning}
\label{cmsr-hyper}

The hyperparameters used in commonsense reasoning experiments for MiLoRA and PiSSA are provided in Tables \ref{tab:pissa-hyper} and \ref{tab:milora-hyper}. The settings for Lily and LoRA using Falcon-Mamba as the backbone are presented in Tables \ref{tab:Lily-cmsr-hyper} and \ref{tab:lora-hyper}.  

Notably, Lily achieves the best performance by adapting only the multi-head self-attention (MHSA) module in LLaMA3-8B, whereas other compared methods adapt all modules, including M$A$. Moreover, Lily utilizes the fewest parameters, demonstrating its superior adaptation capability in low-parameter-budget scenarios.

\begin{table*}[h]
\centering
\caption{Hyperparameter configuration from the MiLoRA paper.}
\begin{tabular}{cc}
\toprule
\multicolumn{2}{c}{\textbf{MiLoRA hyperparameters}} \\
\midrule
Rank r & 32 \\
$\alpha$ of LoRA & 64 \\
$\alpha$ of PiSSA & 32 \\
Dropout & 0.05 \\
Optimizer & AdamW \\
LR & 3e-4 \\
LR Scheduler & Linear \\
Batch Size & 16 \\
Warmup Steps & 100 \\
Epochs & 3 \\
Placement & query, key, value, M$A$ up, M$A$ down \\
\bottomrule
\end{tabular}
\label{tab:milora-hyper}
\end{table*}
\begin{table*}[ht]
\centering
\caption{Hyperparameter configuration from the PiSSA paper.}
\begin{tabular}{cc}
\toprule
\multicolumn{2}{c}{\textbf{PiSSA hyperparameters}} \\
\midrule
$\alpha$ & Same as rank r \\
Dropout & 0.0 \\
Optimizer & AdamW \\
LR & 2e-5 \\
LR Scheduler & cosine \\
Batch Size & 128 \\
Warmup Ratio & 0.03 \\
Epochs & 1 \\
Placement & query, key, value, output, gate, M$A$ up, M$A$ down \\
\bottomrule
\end{tabular}
\label{tab:pissa-hyper}
\end{table*}

\begin{table}[ht]
\centering
\caption{Hyperparameter configuration for LoRA using Falcon-Mamba as backbone.}
\begin{tabular}{cc}
\toprule
\multicolumn{2}{c}{\textbf{LoRA hyperparameters}} \\
\midrule
Rank r & 2 \\
$\alpha$ & 16 \\
Dropout & 0.05 \\
Optimizer & AdamW \\
LR & 3e-4 \\
LR Scheduler & Linear \\
Batch Size & 16 \\
Epochs & 1 \\
Placement & input, delta \\
\bottomrule
\end{tabular}
\label{tab:lora-hyper}
\end{table}

\begin{table*}[ht]
\centering
\caption{Best Hyperparameter configuration for Lily using Falcon-Mamba and LLaMA3 as backbones.}
\begin{tabular}{ccc}
\toprule
 & \textbf{Falcon-Mamba} & \textbf{LLaMA3} \\
\midrule
Rank r & 40 & 16 \\
\textit{ne}\_1 & 4 & 4 \\
\textit{ne}\_2 & 4 & 4 \\
Dropout & 0 & 0 \\
Optimizer & AdamW & AdamW \\
LR & 3e-4 & 3e-4 \\
LR Scheduler & Linear & Linear \\
Batch Size & 16 & 16 \\
Epochs & 1 & 3 \\
Placement & input & query, key, value\\
\bottomrule
\end{tabular}
\label{tab:Lily-cmsr-hyper}
\end{table*}
\begin{table*}[ht]
\centering
\caption{Hyperparameter of Lily on GLUE benchmark.}
\vspace{5pt}
\label{tab:nlu-hyper}
\resizebox{0.9\textwidth}{!}{%
\begin{tabular}{@{}ccccccccc@{}}
\toprule
Hyperparameter & \multicolumn{1}{|c}{STS-B} & \multicolumn{1}{c}{RTE} & \multicolumn{1}{c}{MRPC} & \multicolumn{1}{c}{CoLA} & \multicolumn{1}{c}{SST-2} & \multicolumn{1}{c}{QNLI} & \multicolumn{1}{c}{MNLI} & \multicolumn{1}{c}{QQP} \\ \midrule
\multicolumn{1}{l|}{Optimizer} & \multicolumn{8}{c}{AdamW} \\
\multicolumn{1}{l|}{LR Schedule} & \multicolumn{8}{c}{Linear} \\
\multicolumn{1}{l|}{Learning Rate (Lily)} & \multicolumn{8}{c}{5e-3} \\
\multicolumn{1}{l|}{Learning Rate (Head)} & \multicolumn{8}{c}{5e-3} \\
\multicolumn{1}{l|}{Max Seq. Len} & 512 & 512 & 512 & 512 & 512 & 512 & 512 & 512 \\
\multicolumn{1}{l|}{Lily\_s} & 0.1 & 0.1 & 0.1 & 0.01 & 0.01 & 0.01 & 0 & 0 \\
\multicolumn{1}{l|}{\textit{ne}\_1} & 2 & 3 & 2 & 4 & 2 & 2 & 0 & 0 \\
\multicolumn{1}{l|}{\textit{ne}\_2} & 2 & 3 & 2 & 4 & 2 & 2 & 0 & 0 \\
\multicolumn{1}{l|}{Batch Size} & 64 & 32 & 50 & 64 & 32 & 32 & 0 & 0 \\  \bottomrule
\end{tabular}%
}
\end{table*}
\subsubsection{Natural Language Understanding}
\label{nlu-hyper}

The specific hyperparameter settings for Lily on the GLUE benchmark are provided in Table \ref{tab:nlu-hyper}. We fix the learning rate of both the backbone and the head at $5 \times 10^{-3}$ and instead tune the scaling factor Lily\_s, where Lily\_s $\in \{0.01, 0.1, 1.0\}$. The rank $r$ is fixed at 32, and the random seed is set to 0. The baseline results are taken from FourierFT \citep{gao2024fourierft}.

\begin{table*}[ht]
\centering
\caption{Hyperparameter configuration for Lily on VTAB-1K benchmark.}
\begin{tabular}{ccc}
\toprule
& \textbf{Vision Transformer} & \textbf{Vision Mamba} \\
\midrule
Optimizer & AdamW & AdamW \\
Batch Size & 64 & 64 \\
Learning Rate & 1E-3 & 1E-2\\
Weight Decay & 1E-4 & 1E-3\\
\# Epochs & 100 & 100 \\
LR Decay & cosine & cosine\\
\bottomrule
\end{tabular}
\label{tab:cv-hyper}
\end{table*}

\subsubsection{Visual Task Adaptation Benchmark}
\label{cv-hyper}

We provide the hyperparameters for Lily on the VTAB-1K benchmark in Table \ref{tab:cv-hyper}. Specifically, we fix the learning rate at $1 \times 10^{-3}$ with a weight decay of $1 \times 10^{-4}$. For ViT, we tune the scaling factor Lily\_s $\in \{0.01, 0.1, 1.0, 10.0\}$ to maximize performance, following \citet{jie2023revisiting} and \citet{jie2023fact}. For Vim, we fix Lily\_s to 1.0. Additionally, we search for the hyperparameters \textit{ne}\_1 and \textit{ne}\_2 within the range \{2, 3, 4\}, as these numbers divide the number of layers in the ViT model (12 in ViT-B).  

For Vim, we use the implementation discussed in Section \ref{model}, which does not share $A$s across layers. Therefore, \textit{ne}\_1 in this setting is fixed to the number of layers in Vim (22 in this case), while we search for \textit{ne}\_2 in \{3, 6\} and \{5, 6, 17\} separately for Lily-S and Lily-L. Note that \textit{ne} is only set for the input projection in Vim. For the delta transformation, we use only a single $B$ expert to reduce the parameter cost.  

In the ViT experiments, the rank $r$ is fixed at 16. Meanwhile, in Vim's setting, we tune the ranks $r$ for the delta transformation module and the input projection module separately. We use $(4,4)$ and $(4,8)$ separately for Lily-S and Lily-L.

\subsection{Datasets}
\subsubsection{Commonsense Reasoning}
\label{cmsr-data}
\begin{table*}[h!]
\centering
\caption{Details of the datasets used in our commonsense reasoning tasks.}
\vspace{5pt}
\label{tab:cmsr-data}
\begin{tabular}{@{}llr@{}}
\toprule
\textbf{Benchmark} & \textbf{Description} & \textbf{\# Test Questions} \\ \midrule
ARC-c & Multiple-choice science & 2376 \\
ARC-e & Multiple-choice science & 1172 \\
OBQA & Multi-step reasoning & 500 \\
SIQA & Social implications & 1954 \\
WinoG & Fill-in-a-blank & 1267 \\
PIQA & Physical commonsense & 1830 \\
BoolQ & Yes/no questions & 3270 \\
HellaS & Commonsense NLI & 10042 \\ \bottomrule
\end{tabular}
\end{table*}

We provide a short description of each datasets used in commonsense reasoning experiments in Table \ref{tab:cmsr-data}. 

\subsubsection{Natural Language Understanding}
\label{nlu-data}
We provide detailed information about datasets in the GLUE benchmark in Table \ref{tab:nlu-data}.

\begin{table*}[h!]
\centering
\caption{Information about datasets in the GLUE benchmark, with STS-B being a regression task and all other tasks falling into the categories of single-sentence or sentence-pair classification.}
\vspace{5pt}
\label{tab:nlu-data}
\resizebox{0.9\textwidth}{!}{%
\begin{tabular}{@{}lllllrrr@{}}
\toprule
\multicolumn{1}{l|}{\textbf{Corpus}} & \textbf{Metrics} & \textbf{Task} & \textbf{\# Train} & \textbf{\# Val} & \textbf{\# Test} & \textbf{\# Labels} \\ \midrule
\multicolumn{7}{c}{Single-Sentence Tasks} \\ \midrule
\multicolumn{1}{l|}{CoLA} & Matthews Corr. & Acceptability & 8.55k & 1.04k & 1.06k & 2 \\
\multicolumn{1}{l|}{SST-2} & Accuracy & Sentiment & 67.3k & 872 & 1.82k & 2 \\ \midrule
\multicolumn{7}{c}{Similarity and Paraphrase Tasks} \\ \midrule
\multicolumn{1}{l|}{MRPC} & Accuracy/F1 & Paraphrase & 3.67k & 408 & 1.73k & 2 \\
\multicolumn{1}{l|}{STS-B} & Pearson/Spearman Corr. & Sentence similarity & 5.75k & 1.5k & 1.38k & 1 \\
\multicolumn{1}{l|}{QQP} & Accuracy/F1 & Paraphrase & 364k & 40.4k & 391k & 2 \\ \midrule
\multicolumn{7}{c}{Inference Tasks} \\ \midrule
\multicolumn{1}{l|}{MNLI} & Accuracy & NLI & 393k & 19.65k & 19.65k & 3 \\
\multicolumn{1}{l|}{QNLI} & Accuracy & QA/NLI & 105k & 5.46k & 5.46k & 2 \\
\multicolumn{1}{l|}{RTE} & Accuracy & NLI & 2.49k & 277 & 3k & 2 \\ \bottomrule
\end{tabular}%
}
\end{table*}

\subsubsection{Visual Adaptation Benchmark}
\label{cv-data}
We provide detailed information about all the tasks from VTAB-1K benchmark in Table \ref{tab:vtab-data}.

\begin{table*}[t]
\caption{Detailed information about the datasets in VTAB-1K benchmark.
}
\begin{center}
\scalebox{0.9}{
\begin{tabular}{lllllc}
\toprule[1.5pt]
 & Dataset & Train & Val & Test & \#Classes \\ \midrule
\multirow{19}{*}{VTAB-1k} & CIFAR100 & \multirow{19}{*}{800/1,000} & \multirow{19}{*}{200} & 10,000 & 100 \\
 & Caltech101 & & & 6,084 & 102 \\
 & DTD & & & 1,880 & 47 \\
 & Oxford-Flowers102 & & & 6,149 & 102 \\
 & Oxford-Pets & & & 3,669 & 37 \\
 & SVHN & & & 26,032 & 10 \\
 & Sun397 & & & 21,750 & 397 \\
 & Patch Camelyon & & & 32,768 & 2 \\
 & EuroSAT & & & 5,400 & 10 \\
 & Resisc45 & & & 6,300 & 45 \\
 & Retinopathy & & & 42,670 & 5 \\
 & Clevr/count & & & 15,000 & 8 \\
 & Clevr/distance & & & 15,000 & 6 \\
 & DMLab & & & 22,735 & 6 \\
 & KITTI-Dist & & & 711 & 4 \\
 & dSprites/location & & & 73,728 & 16 \\
 & dSprites/orientation & & & 73,728 & 16 \\
 & SmallNORB/azimuth & & & 12,150 & 18 \\
 & SmallNORB/elevation & & & 12,150 & 18 \\ 
 \bottomrule
\end{tabular}
}
\end{center}
\label{tab:vtab-data}
\end{table*}

\section{Does Sharing $A$s Result in Inferior Performance?}
\label{sharing-lp-analysis}

As mentioned earlier, we adopted a strategy of sharing the $A$ across most of our experiments, ensuring that the number of $A$s and $B$ experts is consistent. This approach offers two key benefits: simplicity and enhanced parameter efficiency. By sharing the $A$, we eliminate the need to set a separate $A$ for each layer, thereby reducing the overall parameter count.

Our decision to share the $A$ is based on the observation of overall redundancy among layers. Specifically, different layers have varying levels of importance \citep{zhang2023adalora}, and some less important layers do not require a dedicated $A$. By not setting a separate $A$ for these layers, we avoid introducing unnecessary parameter overhead while maintaining negligible impact on performance. To test whether sharing $A$ results in inferior performance, we conducted experiments without $A$ sharing on the VTAB-1K benchmark. The results, shown in Table \ref{tab:qv-kv-mlp-analysis}, indicate that the best overall performance ($77.3\%$) is the same as in the $A$-sharing setting. This suggests that even when we employ one $A$ for each layer, the performance gain is negligible, and many of the parameters are, in fact, redundant. However, not sharing $A$s leads to additional parameter overhead, which reduces the parameter efficiency of Lily. Therefore, $A$-sharing is an effective strategy to eliminate redundancy among $A$s and enhance the parameter efficiency of Lily.

\section{Where to Apply Lily in Transformers?}
\label{finetune-transformer-analysis}
\begin{table*}[t]
\caption{Performance on VTAB-1K benchmark when applying Lily to various modules in Transformer. The implementation here does not share $A$ for simplicity (i.e., each layer has one $A$).}
\vspace{-10pt}
\begin{center}
\setlength{\tabcolsep}{0.3pt}
\scalebox{0.75}{
\begin{tabular}{
p{2.30cm}<{}
p{0.75cm}<{\centering}
|
p{0.75cm}<{\centering}
p{0.75cm}<{\centering}
p{0.75cm}<{\centering}
p{0.75cm}<{\centering}
p{0.75cm}<{\centering}
p{0.75cm}<{\centering}
p{0.75cm}<{\centering}
|
p{0.75cm}<{\centering}
p{0.75cm}<{\centering}
p{0.75cm}<{\centering}
p{0.75cm}<{\centering}
|
p{0.75cm}<{\centering}
p{0.75cm}<{\centering}
p{0.75cm}<{\centering}
p{0.75cm}<{\centering}
p{0.75cm}<{\centering}
p{0.75cm}<{\centering}
p{0.75cm}<{\centering}
p{0.75cm}<{\centering}
}
\toprule
\multicolumn{2}{c|}{}&\multicolumn{7}{c|}{\textbf{Natural}}&\multicolumn{4}{c|}{\textbf{Specialized}}&\multicolumn{8}{c}{\textbf{Structured}}\\
&\multicolumn{1}{c|}{\STAB{\rotatebox[origin=c]{90}{Average}}}
&\multicolumn{1}{c}{\STAB{\rotatebox[origin=c]{90}{\textbf{Cifar100}}}}
&\multicolumn{1}{c}{\STAB{\rotatebox[origin=c]{90}{\textbf{Caltech101}}}}
&\multicolumn{1}{c}{\STAB{\rotatebox[origin=c]{90}{\textbf{DTD}}}}
&\multicolumn{1}{c}{\STAB{\rotatebox[origin=c]{90}{\textbf{Flowers102}}}}
&\multicolumn{1}{c}{\STAB{\rotatebox[origin=c]{90}{\textbf{Pets}}}}
&\multicolumn{1}{c}{\STAB{\rotatebox[origin=c]{90}{\textbf{SVHN}}}}
&\multicolumn{1}{c|}{\STAB{\rotatebox[origin=c]{90}{\textbf{Sun397}}}}
&\multicolumn{1}{c}{\STAB{\rotatebox[origin=c]{90}{\textbf{Camelyon}}}}
&\multicolumn{1}{c}{\STAB{\rotatebox[origin=c]{90}{\textbf{EuroSAT}}}}
&\multicolumn{1}{c}{\STAB{\rotatebox[origin=c]{90}{\textbf{Resisc45}}}}
&\multicolumn{1}{c|}{\STAB{\rotatebox[origin=c]{90}{\textbf{Retinopathy}}}}
&\multicolumn{1}{c}{\STAB{\rotatebox[origin=c]{90}{\textbf{Clevr-Count}}}}
&\multicolumn{1}{c}{\STAB{\rotatebox[origin=c]{90}{\textbf{Clevr-Dist}}}}
&\multicolumn{1}{c}{\STAB{\rotatebox[origin=c]{90}{\textbf{DMLab}}}}
&\multicolumn{1}{c}{\STAB{\rotatebox[origin=c]{90}{\textbf{KITTI-Dist}}}}
&\multicolumn{1}{c}{\STAB{\rotatebox[origin=c]{90}{\textbf{dSpr-Loc}}}}
&\multicolumn{1}{c}{\STAB{\rotatebox[origin=c]{90}{\textbf{dSpr-Ori}}}}
&\multicolumn{1}{c}{\STAB{\rotatebox[origin=c]{90}{\textbf{sNORB-Azim}}}}
&\multicolumn{1}{c}{\STAB{\rotatebox[origin=c]{90}{\textbf{sNORB-Ele}}}}\\
\specialrule{0em}{1pt}{1pt}
\hline
\specialrule{0em}{1pt}{1pt}
qv&76.9&73.2&92.3&72.2&99.3&91.4&89.0&56.5&87.6&95.2&84.8&75.9&83.7&65.8&52.8&81.2&87.6&52.4&36.3&43.4 \\
mlp&77.0 & 74.0&92.6&72.2&99.4&91.5&89.0&55.9&88.2&95.5&85.4&76.0&83.3&63.1&53.0&81.4&86.5&53.8&35.6&43.3 \\
qvmlp&77.1& 73.9&93.2&72.7&99.4&91.6&89.7&56.5&87.9&95.3&85.0&76.1&84.6&65.2&53.0&82.1&86.7&53.0&36.0&42.8 \\
kvmlp&77.3&74.0&92.3&72.6&99.3&91.5&89.2&56.7&88.2&95.4&85.3&76.0&84.6&64.9&53.4&81.7&87.5&52.9&36.9&45.2 \\
\bottomrule
\end{tabular}
}
\end{center}
\label{tab:qv-kv-mlp-analysis}
\end{table*}

\section{Performance Analysis on VTAB-1K Benchmark with Lily on Transformer Modules}

PEFT methods have been predominantly explored on the Transformer architecture, which consists of multi-head self-attention (MHSA) and multi-layer perceptron (M$A$) as its core modules. In this section, we analyze the impact of fine-tuned modules on performance using Lily. Specifically, we compare Lily's performance on the VTAB-1K benchmark under four settings:
\begin{itemize}
    \item Applying Lily solely to the query and value transformation module in MHSA (denoted as "qv").
    \item Applying Lily solely to the M$A$ module (denoted as "mlp").
    \item Applying Lily to both the query and value transformation module in MHSA and the M$A$ module (denoted as "qvmlp").
    \item Applying Lily to both the key and value transformation module in MHSA and the M$A$ module (denoted as "kvmlp").
\end{itemize}

To ensure a fair comparison, we tune the hyperparameters to maintain a similar parameter count across all settings. Additionally, to further investigate whether sharing the low-rank projection ($A$) affects performance, we do not share $A$ in this experiment.

The results are presented in Table \ref{tab:qv-kv-mlp-analysis}. We observe that the "kvmlp" setting achieves the best performance, with an average accuracy of $77.3\%$. In contrast, adapting only the MHSA module ("qv") yields the worst performance. Furthermore, we note that adapting both the MHSA and M$A$ modules (qvmlp and kvmlp) generally leads to superior results compared to adapting only one specific module (qv and mlp). This suggests that both M$A$ and MHSA play crucial roles in overall model performance, and adapting both is essential for effective adaptation.

Notably, even when applying Lily solely to the MHSA module, which results in the worst performance among the four settings ($76.9\%$), it still outperforms LoRA by a significant margin ($0.5\%$). This underscores the efficiency of Lily, as it uses fewer parameters than LoRA, even without $A$ sharing.

\section{Where to Apply Lily in Mamba?}
\label{finetune-mamba-analysis}

Nearly all previous PEFT method studies have focused on Transformers, while Mamba is a relatively new architecture, and therefore, there has been little research on PEFT methods for Mamba. In this section, we briefly analyze the pros and cons of adapting Mamba's modules. A Mamba block consists of regular linear projection layers and a core component, the SSM module \citep{gu2023mamba}, \citep{zhu2024visionMamba}. Specifically, in the SSM module, Mamba utilizes parameters ($\Delta$, $A$, $B$, $C$) to transform an input sequence $x(t)$ into an output sequence $y(t)$ using a hidden state $h(t)$. The discretization process converts $A$ and $B$ into $\bar{A}$ and $\bar{B}$, respectively, using the time step size parameter $\Delta$. Structured state space models, inspired by continuous systems, can be computed similarly to RNNs or in the form of global convolution due to their linear time invariance (LTI) property. Mamba introduces a selective property to the structured state space model, tying parameters to the current input. This breaks the LTI property and hinders parallel training. To address this, Mamba employs a hardware-aware algorithm, enabling its SSM module to possess the selective property while performing parallel training. 

To be specific, the discretization process can be expressed as:
\begin{equation}
\label{Eqdiscretization}
\begin{aligned}
    \bar{A} &= \exp(\Delta A) \\
    \bar{B} &= (\Delta A)^{-1}(\exp(\Delta A) - I) \cdot \Delta B
\end{aligned}
\end{equation}
After that, the calculation in Mamba can be expressed as:
\begin{equation}
\label{Eqmamba}
\begin{aligned}
    h_t &= \bar{A} h_{t-1} + \bar{B} x_t \\
    y_t &= C h_t
\end{aligned}
\end{equation}
where $h_t$ is the hidden state at time $t$ and $x_t$ is the corresponding input token. Delta projection is a module in SSM that's learnable and tasked with transforming the parameter $\Delta$. Since adapting the delta projection alone can indirectly adapt the entire SSM module (i.e., $\bar{A}$ and $\bar{B}$ are determined by $\Delta$), it is the most critical component of the SSM module.

We investigate the performance of two adaptation strategies: adapting only the input linear projection layer (denoted as "in") and adapting both the input linear projection layer and the SSM (denoted as "$\Delta$ + in" since we only adapt the delta projection in the SSM module). Our results, as shown in Table \ref{tab:commonsense_mamba}, indicate that applying Lily solely to the input projection yields better performance than applying it to both the input and delta projection modules. This suggests that when adapting Mamba-based models under the paradigm of low-rank adaptation, it is optimal to adapt only the input projection module outside the SSM module. These findings highlight the need for further research into the impact of fine-tuned modules in Mamba on overall performance. Additionally, developing PEFT methods specifically tailored to Mamba-based models, whether for vision or language foundation models, is also a promising direction for future work.

\section{Performance with Different Learning Rates}
\label{lr}

Since we only tuned the learning rate in the commonsense reasoning experiment, we provide the performance of commonsense reasoning under different learning rates in Table \ref{tab:lr-ablation}.

\begin{table*}[t]
\centering
\caption{Commonsense reasoning results of Lily under various leanring rates.}
\resizebox{1.95\columnwidth}{!}
{
\begin{tabular}{lcccccccccc}
\toprule
\textbf{Model}& \textbf{Lr} & \textbf{BoolQ} & \textbf{PIQA} & \textbf{SIQA} & \textbf{HellaSwag} & \textbf{WinoGrande} & \textbf{ARC-e} & \textbf{ARC-c} & \textbf{OBQA} & \textbf{Avg.} \\
\midrule
\multirow{4}{*}{LLaMA3-8B} & 1e-3 & 70.7 & 84.6 & 77.6 & 87.8 & 77.3 & 88.5 & 74.1 & 80.8 & 80.2 \\
& 5e-4 & 71.8 & 86.5 & 77.9 & 82.8 & 83.1 & 88.6 & 76.8 & 81.4 & 81.1 \\
& 3e-4 & 72.9 & 85.6& 77.8 & 92.7 & 83.3 & 89.7 & 77.6 & 82.8 & 82.8 \\

\bottomrule
\end{tabular}
}
\label{tab:lr-ablation}
\end{table*}

\section{Does Selectivity Help?}
\label{off-router}

\begin{figure*}[t]
    \centering
    \includegraphics[width=\linewidth]{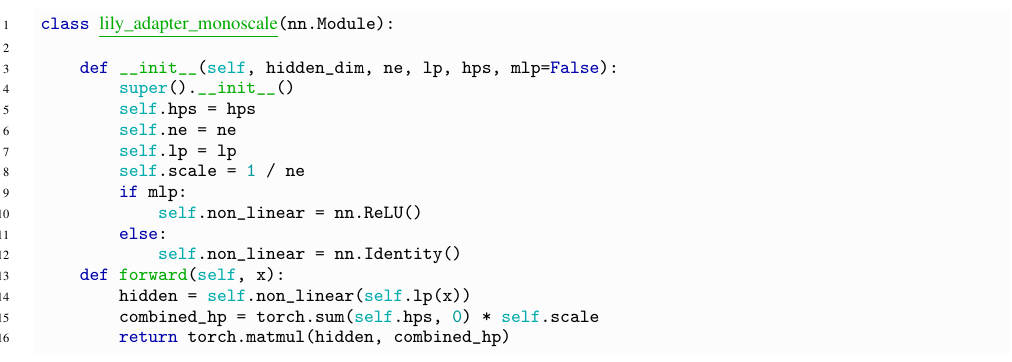}
    \caption{Implementation of Lily with no selectivity.}
    \label{no-router-alg}
\end{figure*}

Lily introduced selective weight combination to selectively incorporate information from other layers. To verify the effectiveness of this selectivity, we remove the router from Lily and evaluate the impact. The modified algorithm without the router is presented in Fig. \ref{no-router-alg}.
We conduct experiments on commonsense reasoning to investigate the effect of removing selectivity from Lily. 

As shown in Table \ref{tab:mono-Lily}, removing selectivity from Lily results in generally poorer performance compared to vanilla Lily. This is likely because the lack of selectivity causes Lily to simply aggregate all the $B$ expert, leading to inferior performance. This validates the design choice of using routers in Lily to selectively allocate weights to $B$ experts, rather than simply summing them.

\begin{table*}[ht]
\centering
\caption{Commonsense reasoning results of Lily without selectivity. We provide results using two learning rates.}
\resizebox{1.95\columnwidth}{!}
{
\begin{tabular}{lcccccccccc}
\toprule
\textbf{Model}& \textbf{Lr} & \textbf{BoolQ} & \textbf{PIQA} & \textbf{SIQA} & \textbf{HellaSwag} & \textbf{WinoGrande} & \textbf{ARC-e} & \textbf{ARC-c} & \textbf{OBQA} & \textbf{Avg.} \\
\midrule
\multirow{3}{*}{LLaMA3-8B} & 3e-4 & 64.0 & 82.6 & 78.5 & 77.0 & 79.6 & 88.4 & 74.5 & 82.0 & 78.3 \\
& 5e-4 & 71.3 & 85.5 & 78.1 & 84.3 & 79.6 & 86.4 & 76.1 & 79.0 & 79.8 \\

\bottomrule
\end{tabular}
}
\label{tab:mono-Lily}
\end{table*}

\section{How to Allocate Parameters?}
\label{keep-params-analysis}

\begin{figure*}
    \centering
    \includegraphics[width=0.75\linewidth]{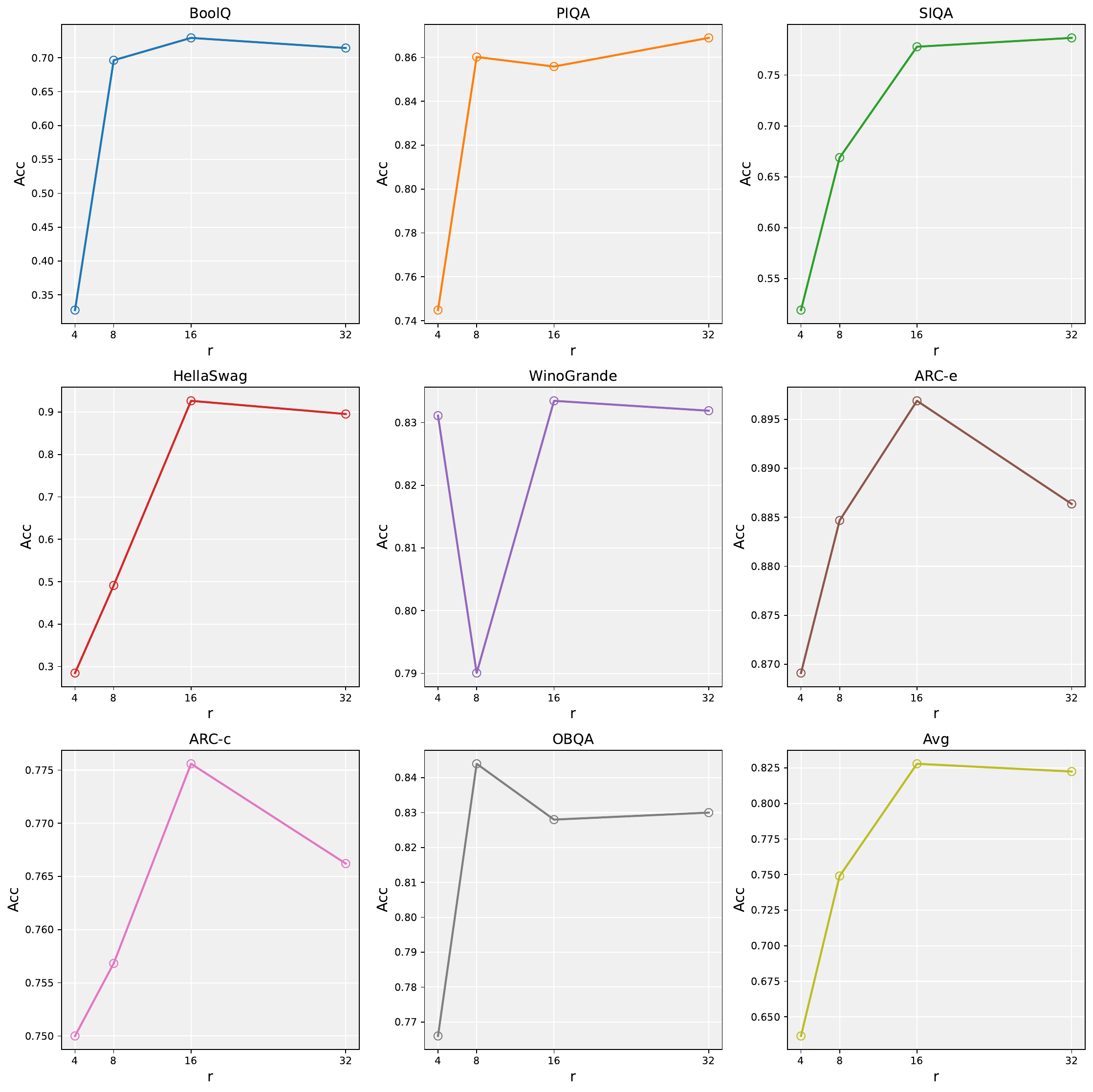}
    \caption{Results on commonsense reasoning tasks when applying different settings of rank. The hyperparameter \textit{ne} is specifically tuned to maintain the same amount of parameter count for a fair comparison.}
    \label{fig:keep-params}
\end{figure*}

Since Lily alters the traditional LoRA's layer-bound setup, increasing the parameters of Lily can be achieved through two approaches: 1) increasing \textit{ne}, i.e., increasing the number of $A$ and $B$ experts, and 2) increasing the rank, i.e., increasing the parameter size of each individual $A$ or $B$ expert. In this section, we investigate which factor has the greatest impact on performance. We conduct experiments on the commonsense reasoning task. Specifically, we maintain the same parameter count and learning rate, and achieve the same parameter count by setting different ranks and adjusting the corresponding \textit{ne} (e.g., r=16, \textit{ne}=4 versus r=8, \textit{ne}=8). The results are shown in Fig. \ref{fig:keep-params}, from which we observe that more $A$ and $B$ experts with smaller rank (i.e., bigger \textit{ne} and smaller rank) generally performs worse. We argue that this is because, although increasing the attention granularity allows for finer details, the resulting performance gain is not as significant as the gain obtained by increasing the rank, i.e., increasing the model's capacity to learn more information. This gives us an insight that, in Lily, increasing \textit{ne} to increase the parameters is less effective than directly increasing the rank in terms of potential performance gain.  
\clearpage

\section{More on Subject-driven Generation}
\label{diffusion-more}

\begin{figure*}[ht]
    \centering
    \includegraphics[width=0.7\linewidth]{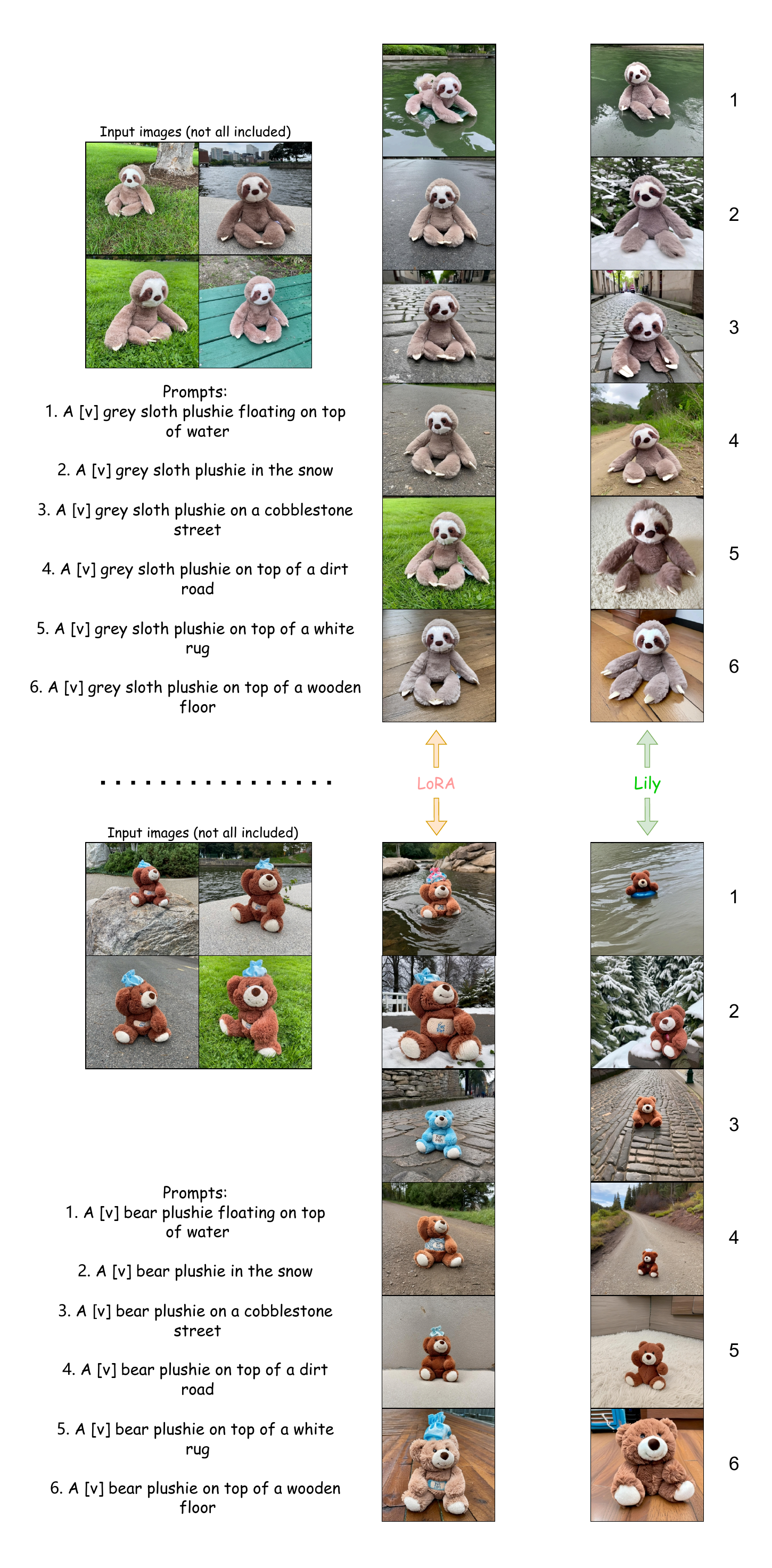}
    \caption{More subject-driven generation results for unreported subjects.}
    \label{fig:more-generation}
\end{figure*}

\begin{figure*}[t]
    \centering
    \includegraphics[width=\linewidth]{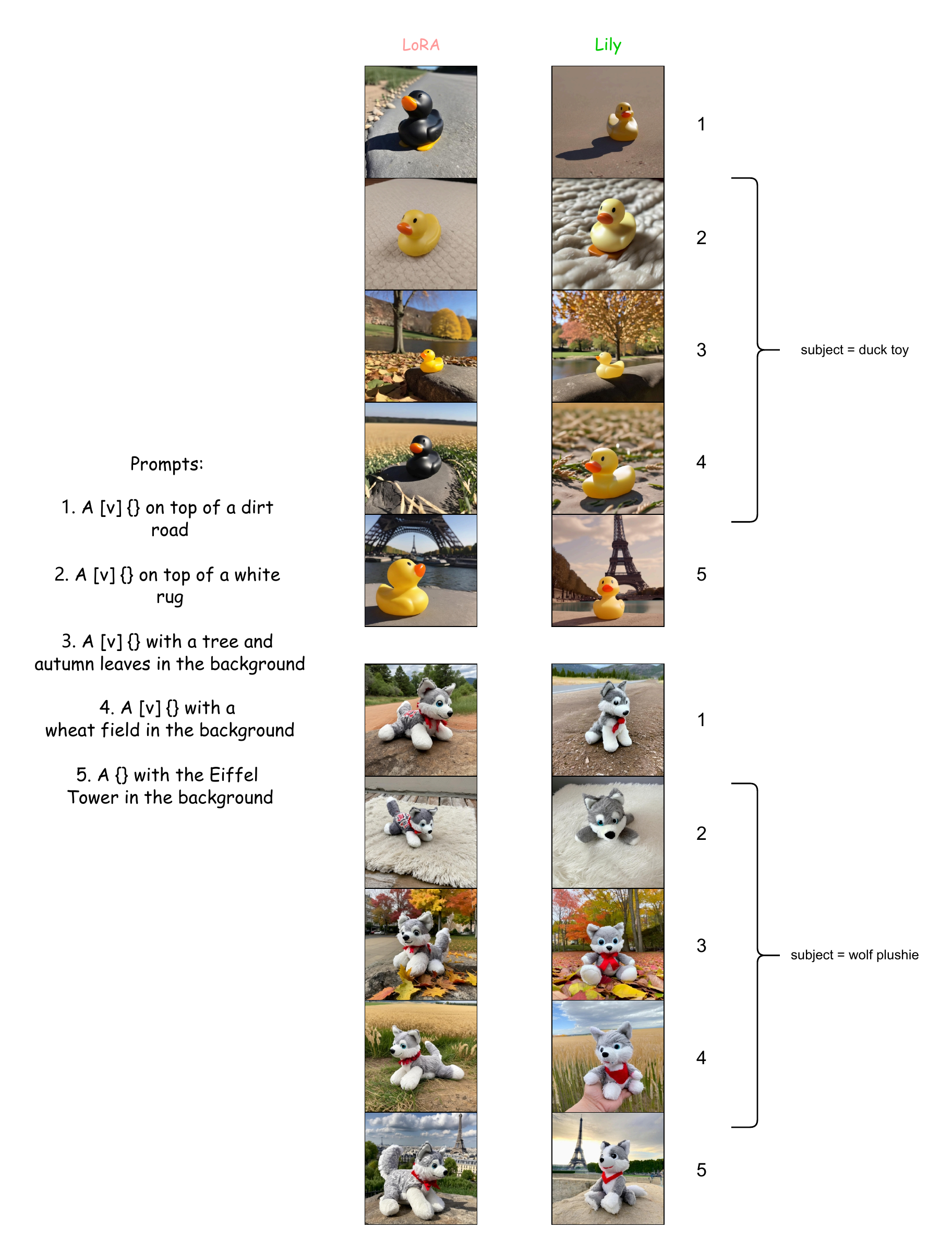}
    \caption{More subject-driven generation results for subjects that are reported in the experiment section.}
    \label{fig:more-generation-2}
\end{figure*}

We provide more results on subject-driven generation in Fig. \ref{fig:more-generation} and Fig. \ref{fig:more-generation-2}.

\section{From a Feature Merging Perspective}

\begin{figure*}[t]
    \centering
    \includegraphics[width=0.99\linewidth]{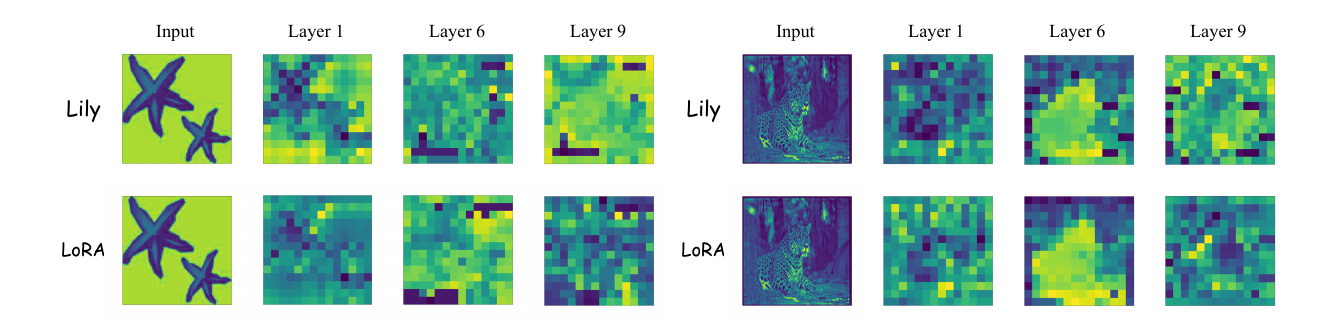}
    \caption{Attention maps of Lily and LoRA. The input images for the example here are taken from Caltech101 datasets from VTAB-1K benchmark. It can be observed that features from a certain layer have more similarity to those in other layers in Lily than in LoRA.}
    \label{fig:attention-map}
\end{figure*}

\begin{figure*}[t]
    \centering
    \includegraphics[width=0.65\linewidth]{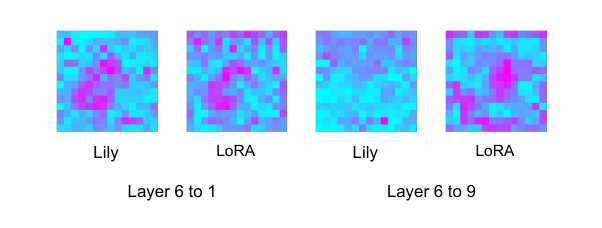}
    \caption{Feature difference measured in absolute distance for each element. We compare Lily and LoRA in terms of the difference between features from different layers. In this example image taken from Caltech101, we visualize the feature difference between layers 6 and 1, as well as between layers 6 and 9.}
    \label{fig:layer-distance}
\end{figure*}

Apart from having higher-rank weight updates than LoRA, Lily also enables comprehensive information access across layers. Lily enables access to information or features from all other layers when adapting a target module at a specific layer thanks to the inter-connectivity of the adapters. We aim to understand how Lily achieves this comprehensive information access from the perspective of visual tasks as shown in Fig. \ref{fig:attention-map}. We can observe that, in Lily, the distinctness of the attention maps between layers is not as pronounced as in LoRA. This validates Lily's ability to enable all-level information access, since adaptation at each layer takes into account features from other layers. Additionally, we specifically visualize the actual feature differences between different layers in Fig. \ref{fig:layer-distance}. We observe that Lily has more points with low feature differences (blue color) than LoRA, indicating that the distinctness of features between layers in Lily is generally lower than in LoRA. This further demonstrates Lily's ability to enable comprehensive information access. Although we enable all-level information access, what prevents the features from becoming completely identical is the selectivity introduced by Lily, which we specify in the following section.

\section{More on Attention Maps of Lily and LoRA}
\label{attention-map-Lily-lora}
\begin{figure*}[t]
    \centering
    \includegraphics[width=0.67\linewidth]{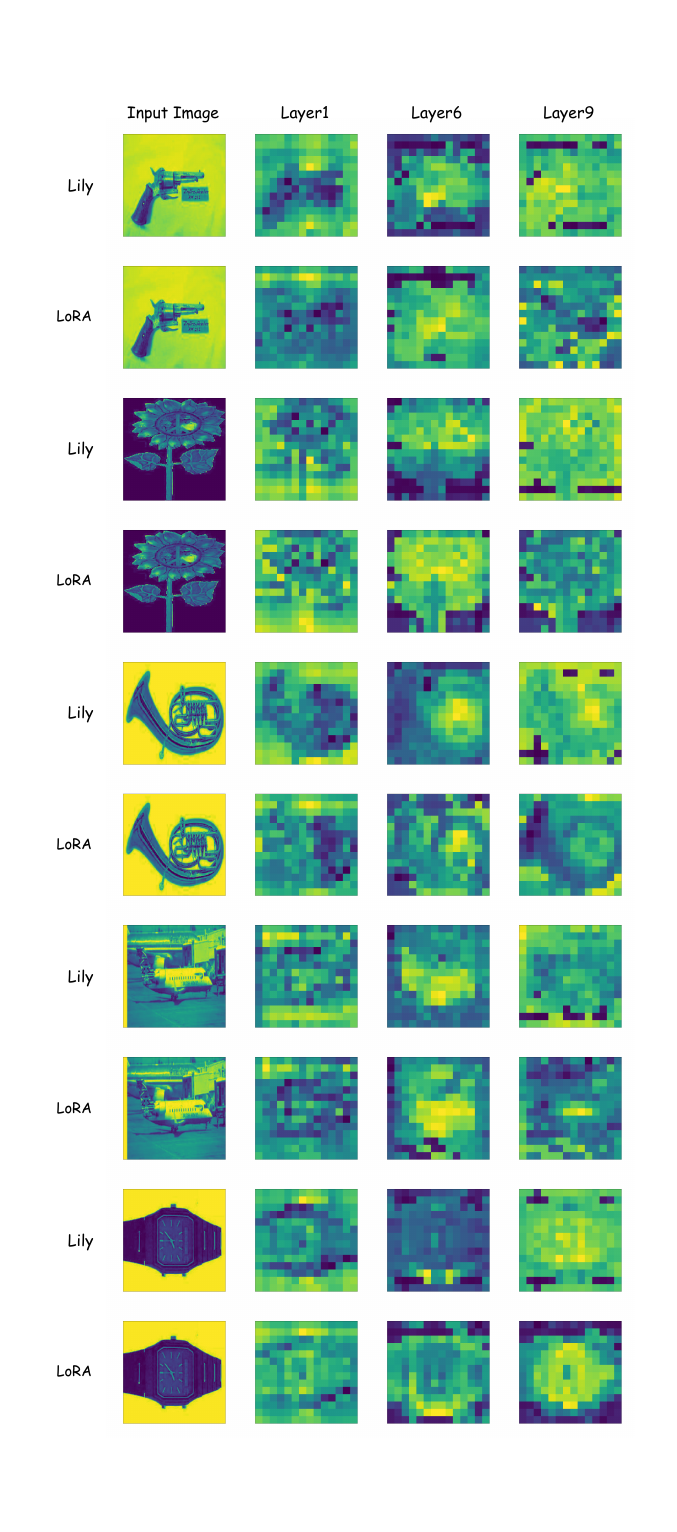}
    \caption{More results of attention maps from LoRA and Lily. All images are taken from Caltech101 dataset.}
    \label{fig:more-attentionmap}
\end{figure*}

We provide more visualization results of the attention map from both LoRA and Lily on Caltech101 dataset from VTAB-1K benchmark in Fig. \ref{fig:more-attentionmap}.

\end{document}